\documentclass[journal]{IEEEtran}

\usepackage{gensymb}
\usepackage{multirow}
\usepackage{color}
\usepackage{cite}
\usepackage{balance}
\usepackage{subfigure}
\usepackage{graphicx}
\usepackage{amsmath,comment}
\usepackage{mathtools}
\usepackage{bbm}

\def \etal {\textit{et al.}}

\ifCLASSINFOpdf
\else
\fi
\usepackage{setspace}

\begin{document}
%
\title{Estimation of Driver's Gaze Region from Head Position and Orientation using Probabilistic Confidence Regions}
%
%
%

\author{Sumit~Jha,~\IEEEmembership{Student Member,~IEEE,}
        Carlos~Busso,~\IEEEmembership{Senior Member,~IEEE,}
\thanks{Sumit Jha and Carlos Busso are with the Erik Johnson School of engineering and Computer Science at the University of Texas at Dallas.}
\thanks{Manuscript received July 03, 2020; revised XXXX XX, 2020.}}

\maketitle

\begin{abstract}
A smart vehicle should be able to understand human behavior and predict their actions to avoid hazardous situations. Specific traits in human behavior can be automatically predicted, which can help the vehicle make decisions, increasing safety. One of the most important aspects pertaining to the driving task is the driver's visual attention. Predicting the driver's visual attention can help a vehicle understand the awareness state of the driver, providing important contextual information. While estimating the exact gaze direction is difficult in the car environment, a coarse estimation of the visual attention can be obtained by tracking the position and orientation of the head. Since the relation between head pose and gaze direction is not one-to-one, this paper proposes a formulation based on probabilistic models to create salient regions describing the visual attention of the driver. The area of the predicted region is small when the model has high confidence on the prediction, which is directly learned from the data. We use \emph{Gaussian process regression} (GPR) to implement the framework, comparing the performance with different regression formulations such as linear regression and neural network based methods. We evaluate these frameworks by studying the tradeoff between spatial resolution and accuracy of the probability map using naturalistic recordings collected with the UTDrive platform. We observe that the GPR method produces the best result creating accurate predictions with localized salient regions. For example, the 95\% confidence region is defined by an area that covers 3.77\% region of a sphere surrounding the driver.
\end{abstract}


%
\IEEEpeerreviewmaketitle

%
%
%
%
\section{Introduction}
\label{sec:introduction}

\IEEEPARstart{R}{oad} safety is a major concern in today's world. The main cause of road accidents is the negligence of distracted drivers \cite{Klauer_2006}. Therefore, monitoring the driver's actions can be useful for predicting their behaviors, creating warnings to avoid impending mistakes due to lack of awareness. Smart vehicles today are equipped with multiple sensors, which provide relevant real-time information inside and outside the vehicle.  The challenge is incorporating heterogeneous information to provide high-level knowledge to understand the driver, the vehicle, and the road. Monitoring the driver's behaviors can also serve as a tool to design advanced user interfaces for infotainment and navigation systems where the drivers naturally interact with the car, without using manual resources \cite{Misu_2015} (e.g., interpreting commands such as ``what is the address of this building?'', while the driver briefly glances towards the target location). With semi-autonomous cars, monitoring the driver behavior can also be helpful in negotiating hand-over control from the vehicle to the driver, or vice-versa.

Visual attention is a major factor when modeling the driver's intentions. The majority of the tasks involved while driving require visual cues. The direction of the driver's gaze strongly depends on the primary driving task and the road condition. Implementing a robust gaze detection system for cars can be helpful in signaling the cognitive state \cite{Li_2015,Koma_2017}, situational awareness \cite{Li_2016,Doshi_2009_2,Wang_2019}, and attention level \cite{Li_2013,Ahlstrom_2013} of the driver. These systems can also be helpful in enhancing in-car dialog systems \cite{Misu_2015}.

In \emph{human-computer interaction} (HCI), the gaze of a subject is estimated by locating the pupil using various appearance based and feature based techniques \cite{Li_2018,Baluja_1994,Jha_2019}. However, these techniques are not practical in a vehicle environment with challenging situations such as varying lighting conditions, high degree of head rotations, and possible occlusions \cite{Jha_2017_2}. Moreover, in the detection of the driver's attention is often more important to achieve robustness across conditions rather than high performance under restricted conditions. A coarse estimation of the driver's visual attention is usually enough for many applications. Following this strategy, studies have proposed the use of head pose to infer the driver's gaze \cite{Tawari_2014_2,Lee_2011_2,Chuang_2014,Rezaei_2014}. Head pose has a strong correlation with gaze, but the relationship is not deterministic \cite{Jha_2016}. Taking the eyes-off-the-road during longer periods significantly increases chances of accidents. Therefore,  drivers tend to have short glances, which involve head and eye movements. This relationship changes according to the driver, primary driving task, secondary driving task, and the traffic condition. Therefore, the head orientation cannot uniquely determine the exact gaze direction.

\begin{figure*}[t]
\subfigure[UTDrive platform]
{
\includegraphics[height=3.5cm]{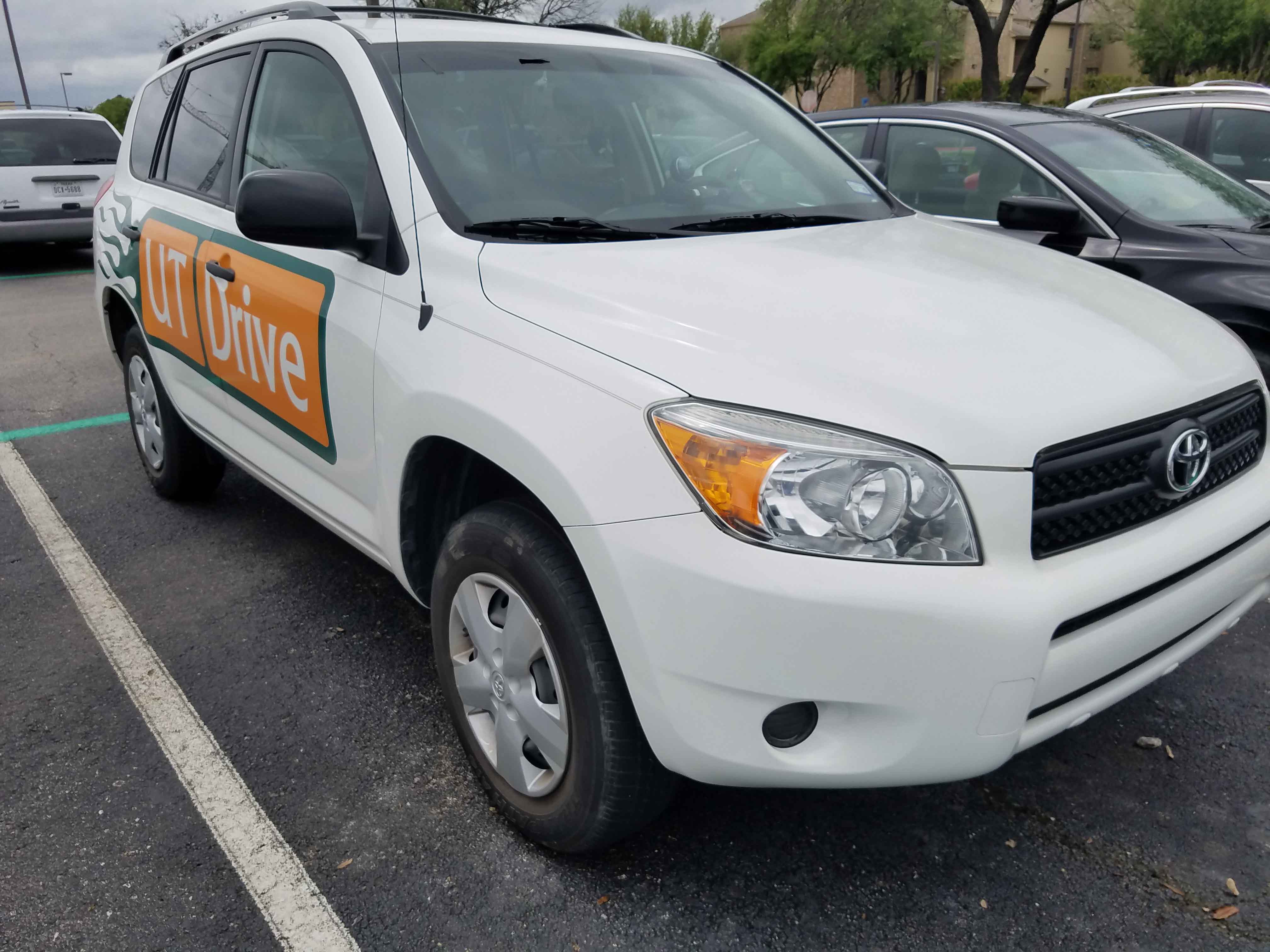}
\label{fig:utdrive}
}
\subfigure[Layout of Markers]
{
\includegraphics[height=3.5cm]{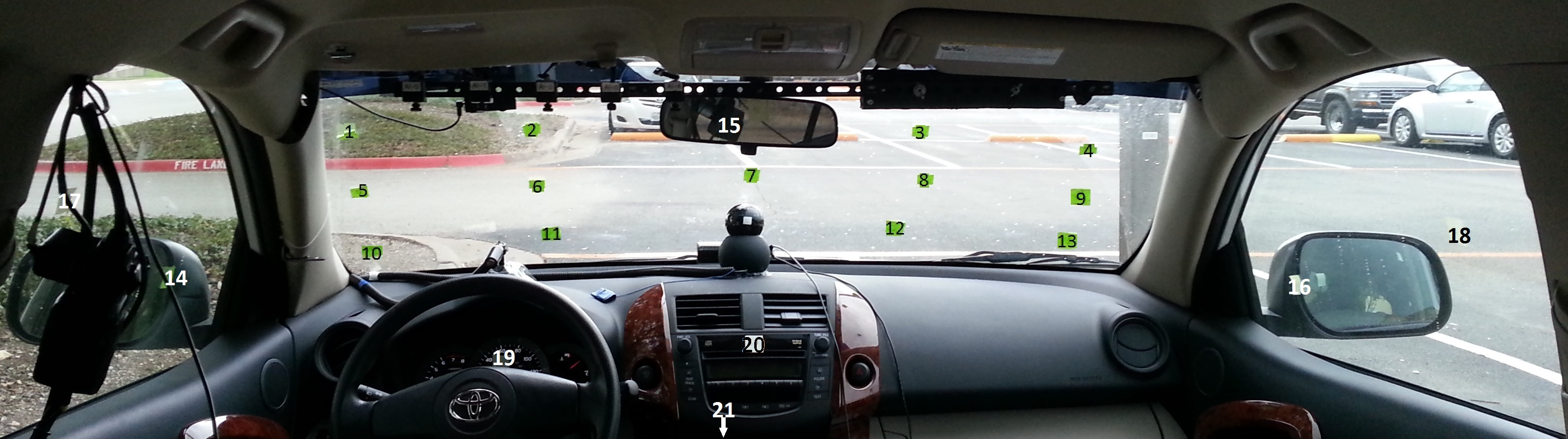}
\label{Fig:markers}
}
\caption{(a) Vehicle used for data collection, and (b) markers placed on the windshield (1-13), mirrors (14-16), side windows (17-18), speedometer panel (19), radio (20), and gear (21). The subjects were asked to look at these markers.}  
\label{fig:sameLabels}
\end{figure*}

Instead of aiming to detect the precise gaze direction, this paper proposes to estimate a probabilistic visual map describing the region of visual attention where the driver is most likely to direct his/her gaze. Building upon our previous work \cite{Jha_2017,Jha_2018_2}, we propose to create this probabilistic visual map using a two dimensional Gaussian distribution that is directly learned from data. The formulation relies on \emph{Gaussian process regression} (GPR) to predict the distribution of the gaze given a certain position and orientation of the driver's head. The proposed model provides not only the probabilistic visual map, but also confidence regions, which can be extremely useful for HCI applications for infotainment and navigation systems, and \emph{advanced driver assistance systems} (ADAS). The size of the salient region decreases when the confidence of the model increases, learning all the parameters of the models directly from the data. We train and evaluate the system with recordings from real driving scenarios using affordable equipment that can be easily installed on regular cars.

The experimental evaluation demonstrates the effectiveness of the proposed GPR system, analyzing the tradeoff between accuracy and spatial resolution of the probabilistic visual map. We compare our proposed solution with alternative machine learning methods to predict the visual maps, including simple regression techniques, deep neural networks and \emph{mixture density networks} (MDNs). The results indicate 
that our proposed model offers the best accuracy and spatial resolution in predicting the probabilistic region of the driver's gaze. For example, 95\% of the target markers lie inside the probabilistic region predicted by the system, where its temporal resolution includes 3.77\% of a sphere surrounding the user's range of vision. Finally, we demonstrate the benefit of the proposed probabilistic model by mapping the probabilistic visual map to areas on the road, allowing us to identify coarse regions outside the car that the driver is directing his/her gaze to.

This study is organized as follows. Section \ref{sec:relatedWork} discusses related studies about the importance of visual attention when studying driver's behavior. Section \ref{sec:data} describes the data collection procedure that we followed to train and evaluate our algorithms. Section \ref{sec:methodology} describes the proposed method to obtain the probabilistic salient visual map to represent visual attention. It also introduces the baseline methods. Section \ref{sec:Results} discusses the results obtained from different models, comparing the tradeoff between spatial resolution and accuracy of the probabilistic salient visual maps. Finally, Section \ref{sec:concl} concludes the study, suggesting future research directions.


\section{Related Work}
\label{sec:relatedWork}
\subsection{Visual Attention of the Driver}
\label{ssec:relvis}

Maintaining visual attention while driving a vehicle is important to reduce hazard scenarios. Drivers obtain most information through vision, which is important to maintain road awareness and to complete driving maneuvers \cite{Li_2016}. Therefore, several studies have considered the visual patterns of the driver, creating useful automatic tools for intelligent vehicle systems. Liang and Lee \cite{Liang_2010} conducted experiments by inducing visual distraction, cognitive distraction and a combination of both by asking subjects to perform distracting tasks while operating a driving simulator. They observed that the driving performance was worse when the subjects were performing visually distracting tasks compared to the performance when performing a combination of visual and cognitive distracting tasks. 

Robinson \etal \cite{Robinson_1972} studied the visual search patterns of a driver by looking at his/her head movements during lane changes and when entering a highway after a stop sign. They observed longer search times at a stop sign, where the drivers had to observe the whole scene before making a decision. In contrast, for lane change actions the search time was shorter since the driver had to make quick decisions. Underwood \etal \cite{Underwood_2003} used eye trackers to study the eye movement behavior of experienced and novice drivers in three different types of roads: rural, suburban and divided highways. They analyzed the most common sequences of fixation in various regions of the road to compare the driving behavior. They observed that a novice driver tends to change his/her fixation more often, while an experienced driver tends to use peripheral vision to pick up subtle information such as the demarcation of lanes. 

Understanding visual attention can also help us infer information about visual and cognitive distractions. Sodhi \etal \cite{Sodhi_2002} used a head-mounted, eye-tracker in a vehicle, where they asked multiple subjects to drive a predetermined route while performing tasks that stimulate distractions. They used the eye-tracker to obtain the position and diameter of the pupil. They studied the impact of infotainment systems on driving by stimulating various cognitive and visual distractions. The study observed that the eye movement patterns changed when the driver was distracted by a secondary task. Kutila \etal \cite{Kutila_2007} recorded the face of the driver with stereo cameras in a naturalistic driving scenario. They used head and gaze information along with lane position and \emph{controller area network} (CAN)-Bus data to detect visual and cognitive distraction. Gaze data was obtained using a gaze tracker. The driver's visual attention is inferred using eyes-off-the-road duration. The eye movement is fused with cognitive workload inferred from the driving data to obtain cognitive distractions. Liang \etal \cite{Liang_2007} designed a \emph{support vector machine} (SVM) classifier that used measures of driving performance such as steering angle and lane position, and features from eye movement data such as fixations and saccades to detect cognitive distraction. They obtained an accuracy of 96.1\% in a simulated environment. Murphy \etal \cite{Murphy_2008} implemented a real-time system to track the six degrees of freedom of the head pose of a driver. They designed an appearance based particle filter to design a 3D model of the face in augmented reality. Rezaei and Klette \cite{Rezaei_2014} monitored both the driver and the road to find possible hazard situations. They designed an asymmetric \emph{active appearance model} (AAM) to predict the driver's head pose, which was used in conjunction with features extracted from the vehicles detected on the road to design a fuzzy logic based system to predict the risk level of the driving situation.

Understanding the driver's behavior is even more relevant with the advances in autonomous cars. Information and datasets derived from drivers in naturalistic conditions, including their visual attention, can be instrumental in the design of autonomous cars \cite{Alletto_2016}, following the ideas of behavior cloning \cite{Bojarski_2016}. Likewise, cars that are aware of the driver's visual attention can more effectively negotiate hand over situations. Zeeb \etal \cite{Zeeb_2016} compared the take-over time and quality between a distracted driver and an attentive driver. The drivers were asked to be involved in secondary tasks such as watching videos and writing emails. They observed that while a driver could quickly resume control of the car when prompted, the quality of the take over was worse when the driver was distracted.

\subsection{Prediction of Visual Attention}
\label{ssec:predictionVisual}

Several studies have worked on predicting the visual attention of the driver, realizing the importance of visual attention in monitoring the behaviors of the driver. The most common approach is to partition the gaze region of the driver into different gaze zones. Then, the problem is formulated as a classification problem to identify the area that the driver is directing his/her attention. Tawari and Trivedi \cite{Tawari_2014_2} video recorded the face of the driver from two different angles in the car to capture the head pose. The two cameras increased the angular range of the head pose estimation. The task was to classify the driver's gaze into eight different gaze zones. They used annotations obtained from human experts as the ground truth for the target gaze zone, training a random forest classifier with the head pose as features. The zone predictions had high confusion between adjacent zones such as looking forward and looking at the speedometer. Lee \etal \cite{Lee_2011_2} suggested a robust method to predict the yaw and pitch of the head. The method relied on simple edge features from the face, making the approach robust to rotation and illumination, and fast enough to be run in real time. They used the predicted yaw and pitch angles to identify one of the 18 predefined gaze zones using an SVM classifier. Chuang \etal \cite{Chuang_2014} designed a gaze estimation system using a smartphone camera. They placed the smartphone on the dashboard to record the driver's face. They used the location of the eyes, nose and mouth regions as features to classify the gaze among eight different zones. Vora \etal \cite{Vora_2017} tried a generalized approach to classify gaze zones which is subject invariant. They used \emph{convolutional neural networks} (CNNs) to obtain the gaze zone from the driver's facial image. The best network achieved a 93.36\% accuracy while performing a seven class classification task (six gaze zones plus a class for eye closure)

While the gaze zone provides useful information about the visual attention of the driver, this information is too coarse for several applications. However, it is challenging to design gaze estimation methods with high precision that work well inside a vehicle. Although there is a strong relationship between head movement and gaze direction, the relation is not one-to-one \cite{Jha_2016}. In naturalistic driving scenarios, the driver relies not only on head movements to direct his/her gaze toward a target location, but also on eye movement. The interplay between head and eye movements depends on the cognitive load of the driver and the underlying driving task. A feasible alternative to gaze zone estimation or unreliable gaze algorithms that do not work in a vehicle is the definition of a probabilistic salient visual map describing the visual attention of the driver. This probabilistic salient visual map can be used to define spatial confidence regions describing the direction of the driver's gaze. This study pursues this novel formulation, creating models that capture the relationship between head pose and gaze, creating a probability distribution of the gaze given the orientation and position of the driver's head.

\begin{figure}[t]
  \centerline{\includegraphics[width=8cm]{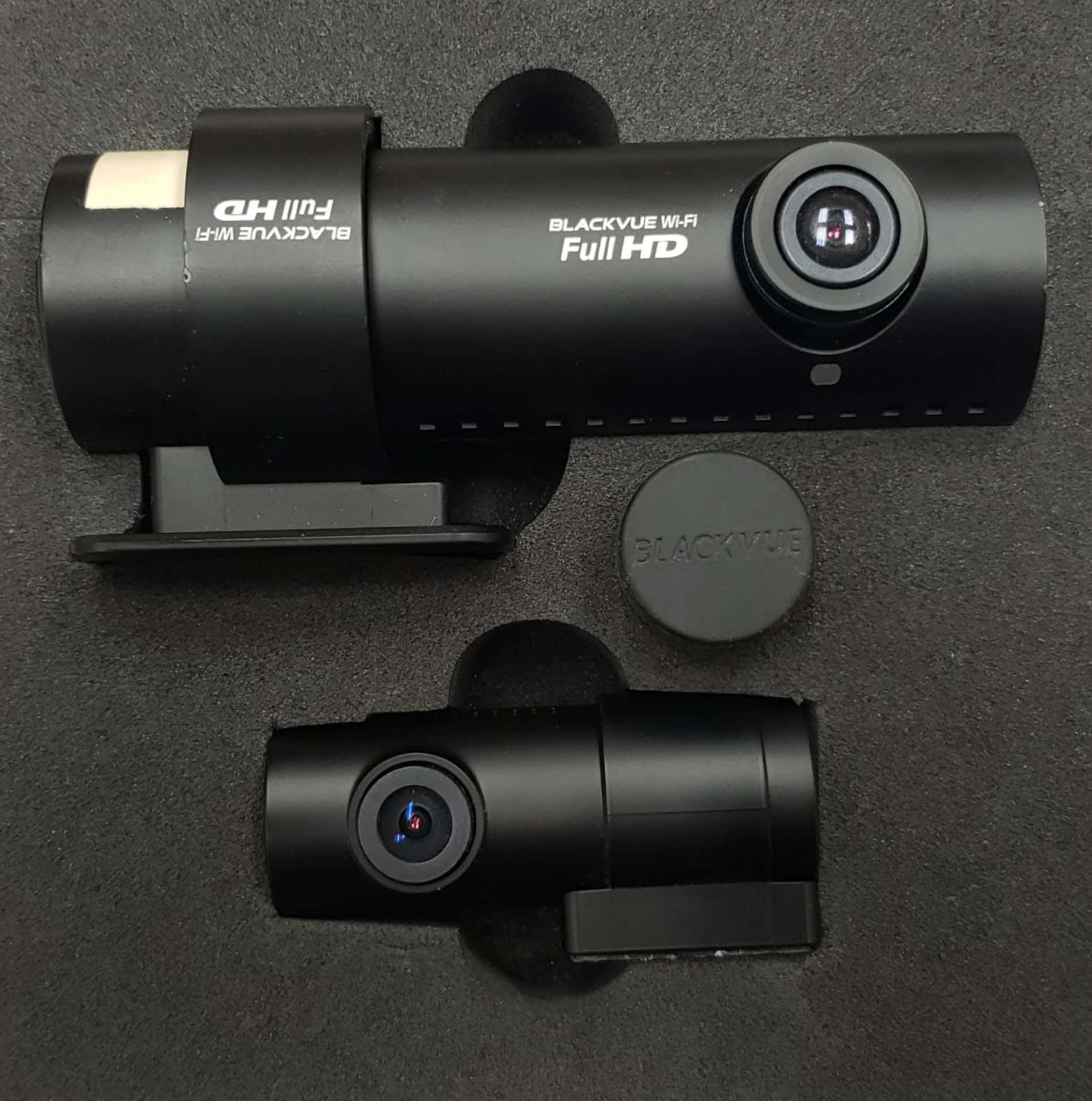}}
\caption{Dash camera (Blackvue DR-650GW-2ch) used to record the face of the driver (primary camera) and the road (secondary camera).}
\label{Fig:blackvue}
\end{figure}

\section{Data Collection}
\label{sec:data}
This study uses recordings from real driving scenarios collected with the UTDrive platform \cite{Angkititrakul_2007_2,Angkititrakul_2007}, which is a vehicle equipped with multiple sensors (Fig. \ref{fig:utdrive}). The UTDrive has been successfully used to study driver behaviors \cite{Hansen_2017,Jain_2011,Li_2013_2}. Instead of using the specific sensors from this car, we decided to only use the commercially available dash camera Blackvue DR-650GW-2ch (Fig. \ref{Fig:blackvue}), which can be easily installed in any regular car. The device features two cameras along with a \emph{global positioning system} (GPS) and accelerometer sensors. The front camera was used to record the road view, while the rear camera was used to record the face of the driver. The system is currently implemented offline. However, the dash camera has WiFi connection which can be utilized for real-time implementations. This setting is ideal for in-vehicle solutions in all cars, regardless of their proprietary built-in sensors.

\subsection{Data Collection Protocol}
\label{ssec:Protocol}

For the analysis, we require data where we know the ground truth information about the direction of the gaze. We achieve this goal by asking the driver to look at predefined markers. We place 21 numbered markers on the windshield (\#1-\#13), mirrors (\#14-\#16), side windows (\#17-\#18), speedometer panel (\#19), radio (\#20), and gear (\#21) (Fig. \ref{Fig:markers}). Then, we ask the subjects to look at these markers multiple times, where we carefully annotated the corresponding timing information. We recruited 16 students (10 males, 6 females) with valid US driver licenses from the University of Texas at Dallas. We designed a three-phase protocol:

\begin{figure}[t]
  \centerline{\includegraphics[width=8cm]{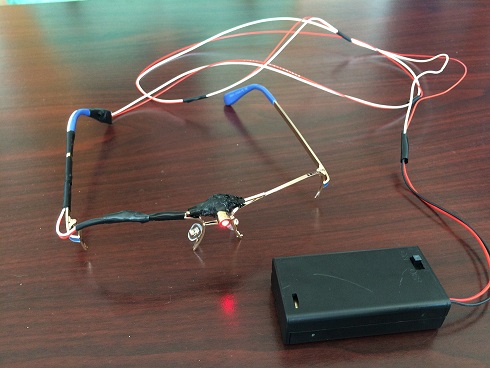}}
\caption{Laser pointer mounted on a glass frame for the controlled head pose condition during phase 3 of the data collection.}
\label{fig:laser}
\end{figure} 

\noindent
\textbf{Phase 1:}
The first phase is recorded when the vehicle is parked. The subject is asked to sit in the driver seat, looking at different markers. The numbers are called out in random order and the driver is asked to look at the corresponding points. Each number was repeated five times in random order. We did not provide any further instruction. The goal of this phase is to estimate and model the gaze-head relationship when our subjects are not driving. They have plenty of time to complete this task without worrying about visual, manual and cognitive demands associated with the driving tasks. The drivers can also get familiar with the task in a safe environment.

\noindent
\textbf{Phase 2:} 
The second phase consists of the same task while the subject is driving the vehicle. The subject is asked to drive on a straight road with low traffic. A passenger reads the numbers, pointing to the target location reducing the cognitive demand of the task. The numbers are requested only when the driver does not have to perform any maneuver. The safety of the subject is our first priority. We do not provide any additional instruction on how to look at the markers. We use this phase to estimate and model the gaze-head relationship while the subject is driving.

\noindent
\textbf{Phase 3:}
During the third phase, we ask the driver to park the car and perform the same task again. This time, the driver is asked to look at each marker directing his/her head toward the point. In this controlled condition, the gaze of the driver is the same as the head pose, without the bias added by the movement of the eye. To enforce this requirement, we request the driver to wear a glass frame with a laser mounted at the center (Fig. \ref{fig:laser}). The driver is asked to point the laser towards the marker. Each number was repeated three times at random for each marker. This phase provides valuable data, where the gaze is exactly aligned with the head orientation. 

Additionally, we asked the last three of our subjects to look at specific locations on the road including billboards, street signals, and buildings to validate our systems in real-world applications. This data is used to assess the mapping between gaze detection and objects on the roads.

\subsection{Head Pose Estimation Using AprilTags}
\label{ssec:AprilTags}

\begin{figure}[!t]
\centering
\subfigure[AprilTag]
{
\includegraphics[height=2.6cm]{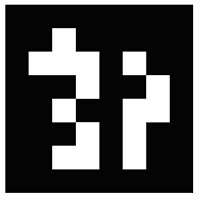}
\label{fig:apriltag_sample}
}\hspace{-0.2cm}
\subfigure[Headband with AprilTags]
{
\includegraphics[height=2.6cm]{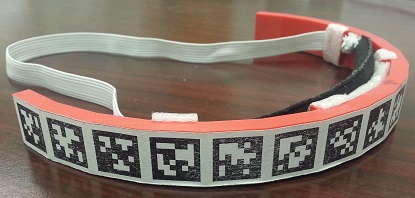}
\label{fig:apriltag_Headband}
}
\caption{(a) Example of a AprilTag, (b) Headband with AprilTags for robust head pose estimation.}
\label{fig:apriltag}
\end{figure}

It is challenging to use computer vision algorithms in a car environment. In our previous work \cite{Jha_2017_2}, we demonstrated that the robustness of a state-of-the-art head pose estimation algorithm was low for non-frontal faces rotated more than 45\degree. For this analysis, we aim to have more robust estimations of head poses regardless of the head orientation. We achieve this goal by using a headband with AprilTags (Fig. \ref{fig:apriltag_sample}). For future work, we can rely on depth cameras to obtain robust head pose estimation \cite{Hu_2020}.

\begin{figure*}[t]
\centering
\subfigure[face camera]
{
\includegraphics[width=5.7cm]{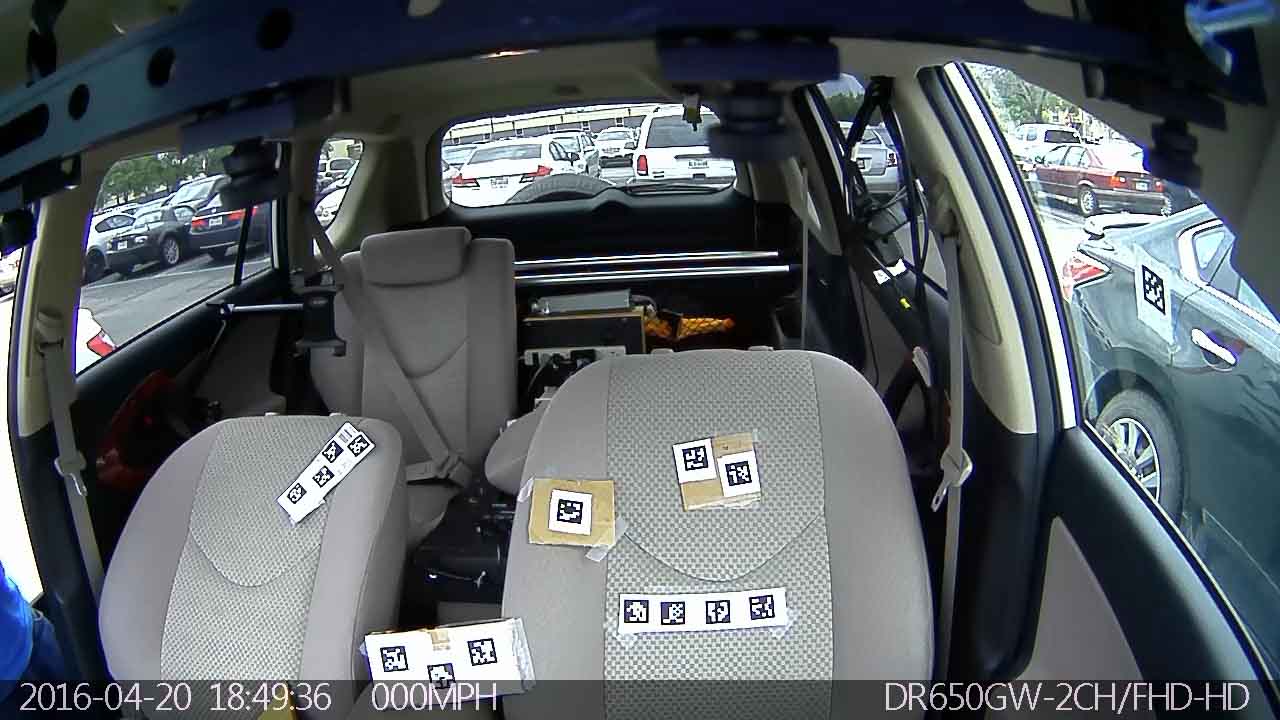}
\label{fig:ref_markers_a}
}\hspace{-0.2cm}
\subfigure[calibration camera, view 1]
{
\includegraphics[width=5.7cm]{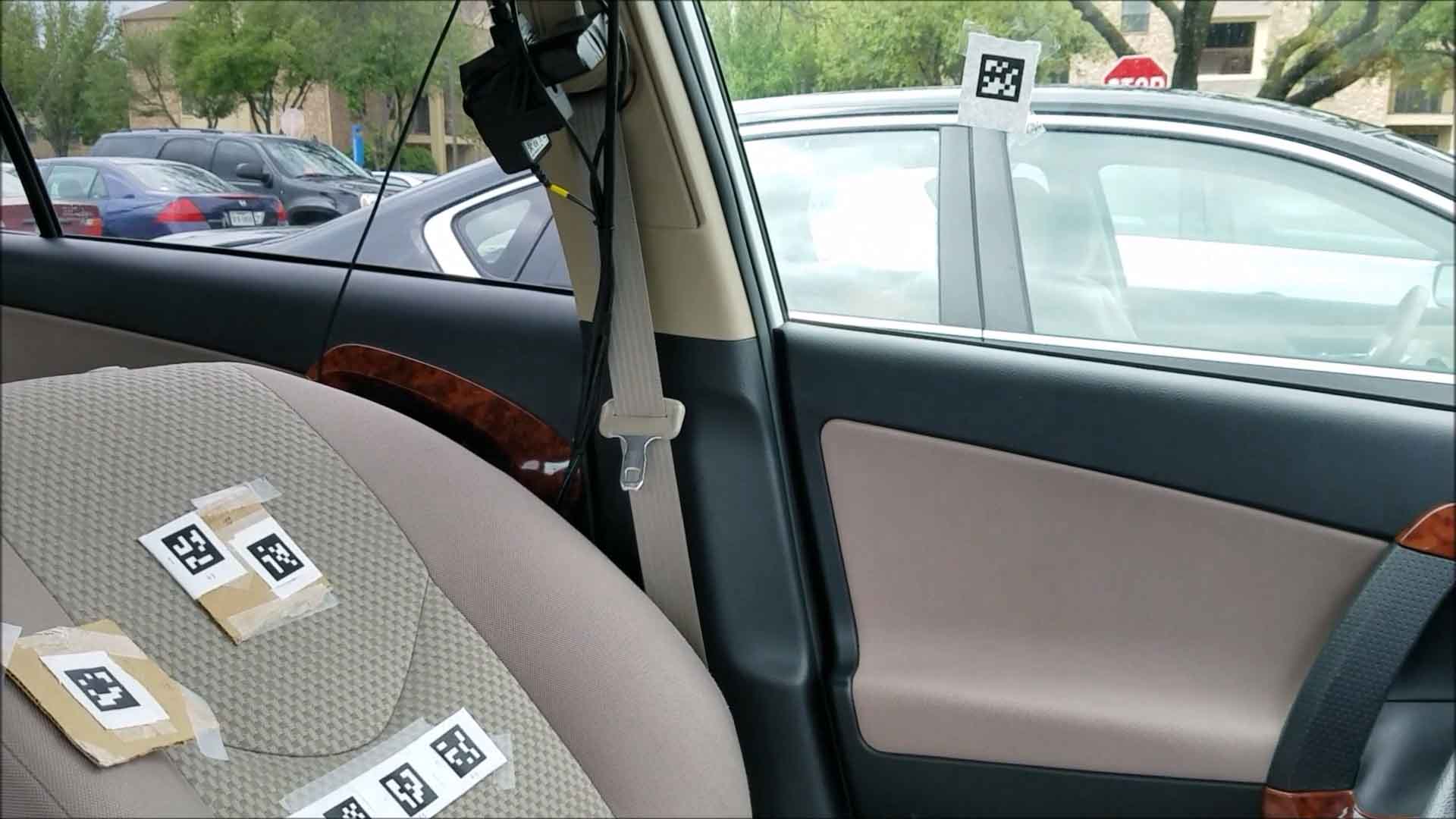}
\label{fig:ref_markers_b}
}\hspace{-0.2cm}
\subfigure[calibration camera, view 2]
{
\includegraphics[width=5.7cm]{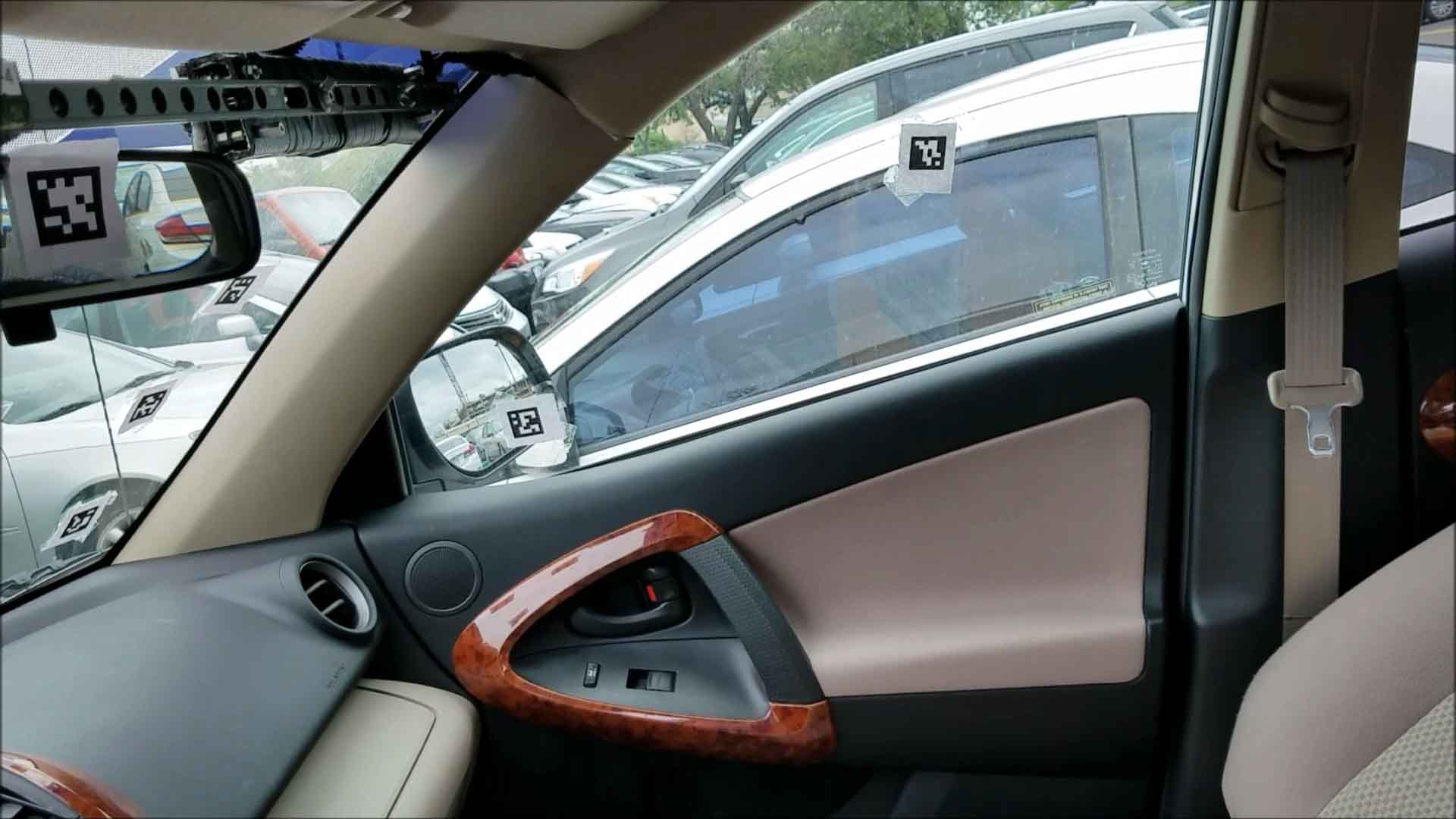}
\label{fig:ref_markers_c}
}\\
\subfigure[calibration camera, view 3]
{
\includegraphics[width=5.7cm]{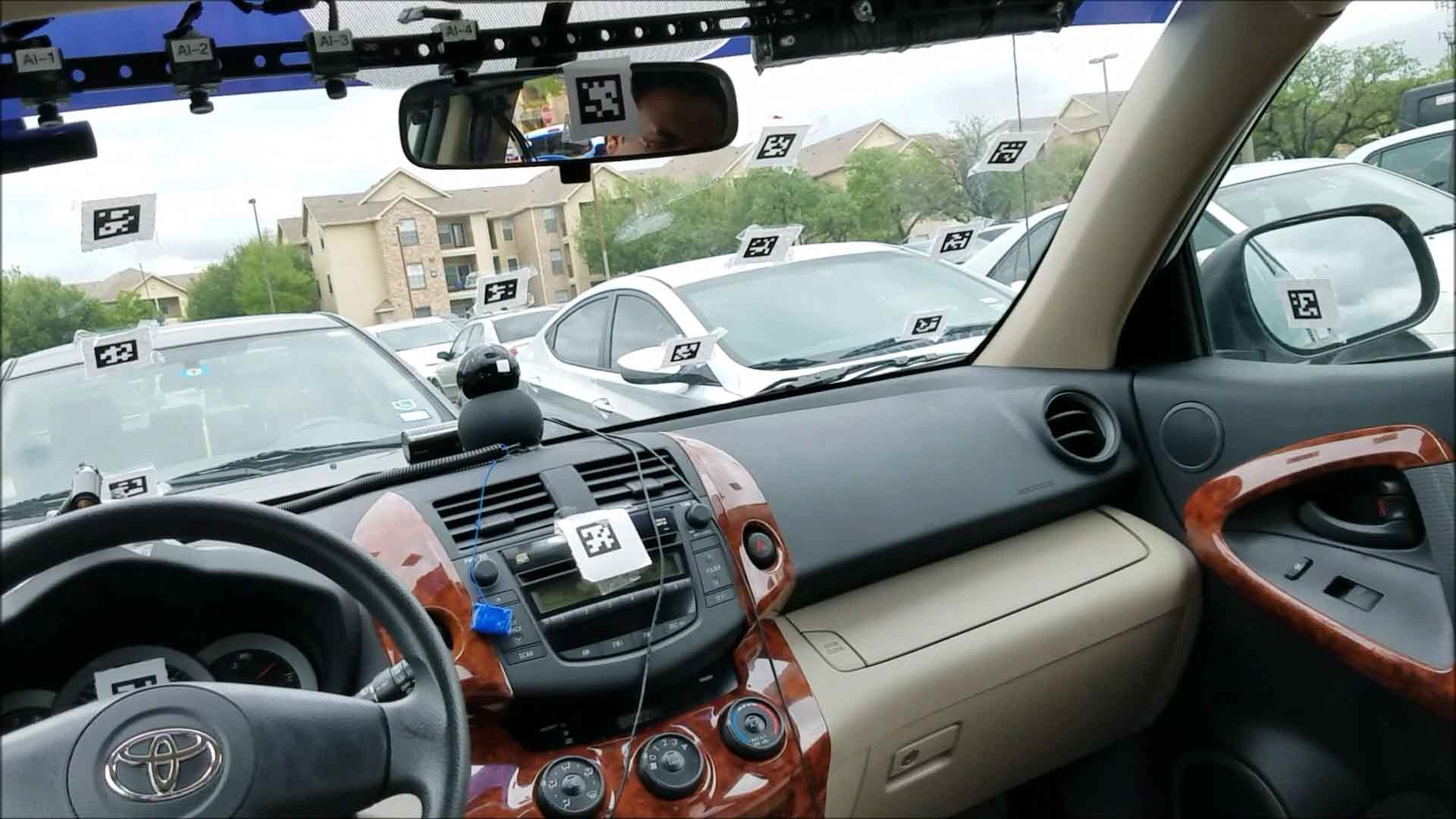}
\label{fig:ref_markers_d}
}\hspace{-0.2cm}
\subfigure[calibration camera, view 4]
{
\includegraphics[width=5.7cm]{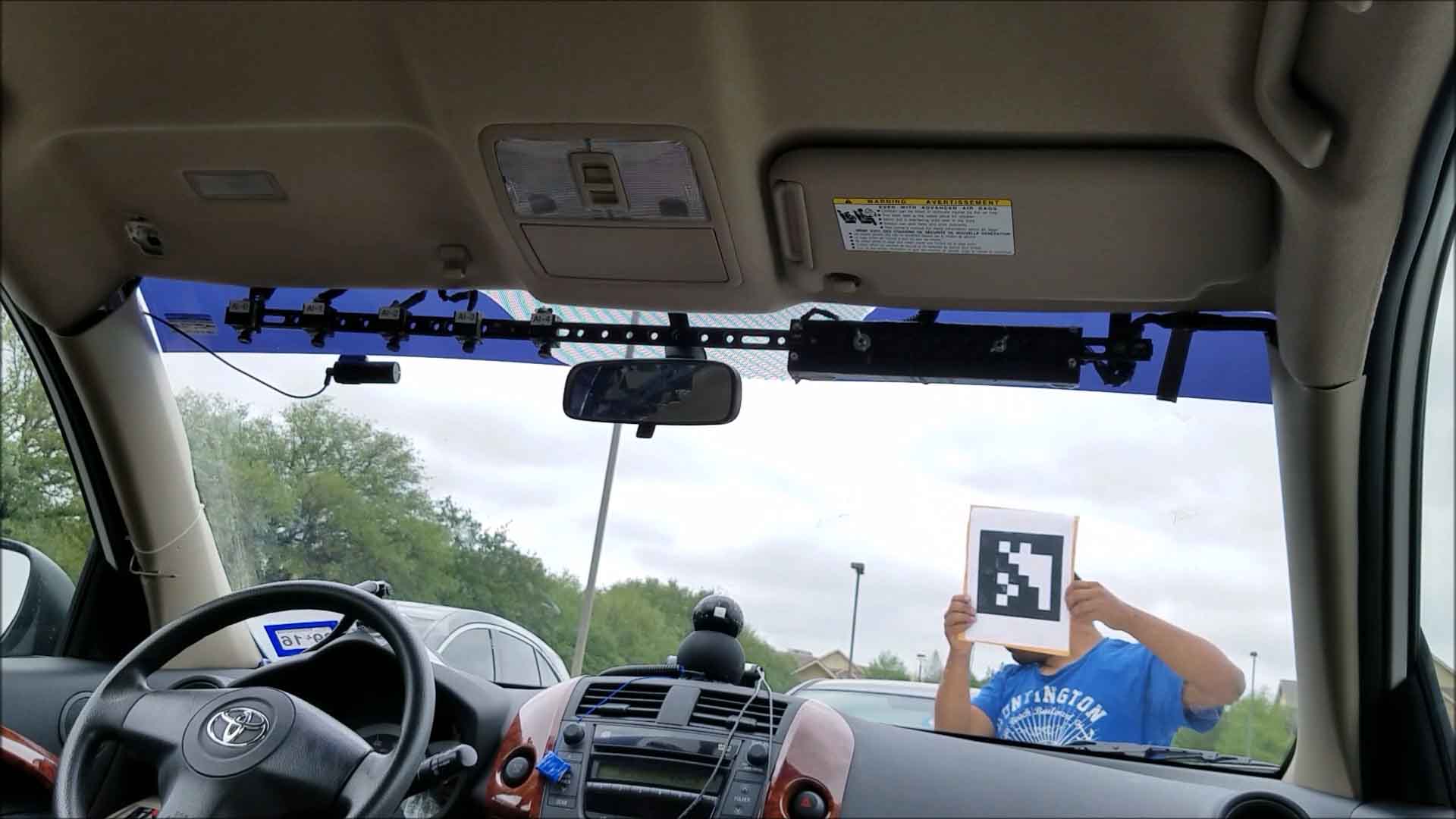}
\label{fig:ref_markers_e}
}\hspace{-0.2cm}
\subfigure[Road camera]
{
\includegraphics[width=5.7cm]{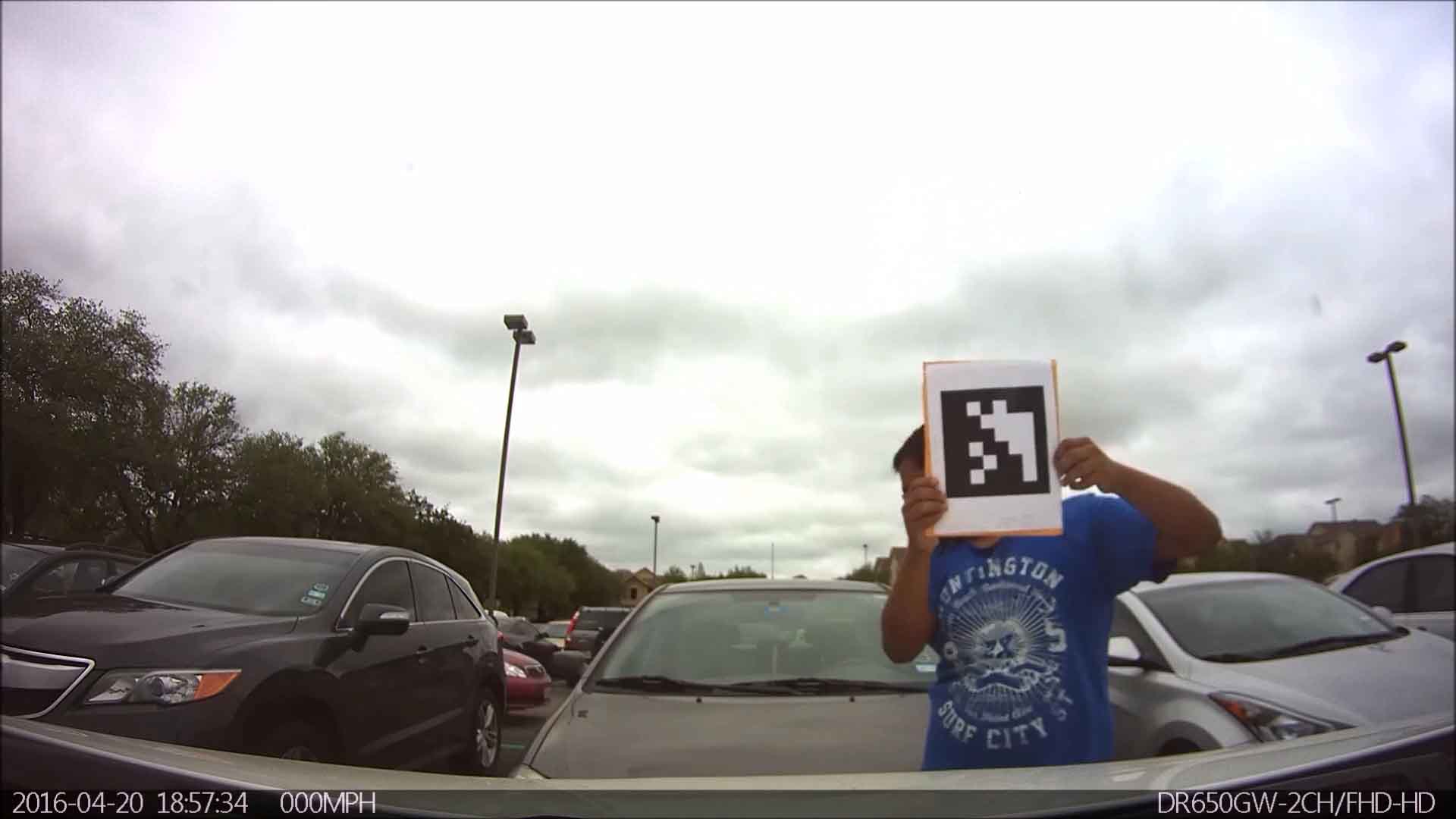}
\label{fig:ref_markers_f}
}
\caption{Description of the process to calibration the cameras and define a common reference system. Figures (a)-(d) show the calibration for the markers' locations and the face camera. Figures (e)-(f) show the calibration for the road camera.}
\label{fig:ref_markers}
\end{figure*}

AprilTags \cite{olson_2011} are 2D barcodes primarily used for augmented reality applications, robotics and camera calibration. Figure \ref{fig:apriltag_sample} shows an example of an AprilTag. The unique black and white patterns of each AprilTag are easy to automatically detect with computer vision algorithms. From the pattern, it is possible to accurately estimate the position and orientation of the tags. Instead of using a single AprilTag, we rely on many tags to robustly estimate the position and orientation of the head. For this purpose, we designed a headband with 17 square faces (2 $\times$ 2 cm each), separated by an angle of 12\degree. Each of the square faces contains a 1.6 cm $\times$ 1.6 cm unique tag. Figure \ref{fig:apriltag_Headband} shows the headband worn by the participants. During the data collection, the subject is asked to wear the band for the entire recording. The selected design allows us to observe multiple tags for each video frame, regardless of the orientation of the driver's head. Therefore, we can robustly infer the position and orientation of the headband. 

The AprilTags from the headband are used to obtain the position and orientation of the driver's head. The AprilTag toolkit provides an estimate of the position and orientation of each tag present in an image. The structure of the headband and orientation of the visible bands help us estimate the pose of the headband.

The use of the headband facilitates the analysis of head pose regardless of the orientation of the head or the environmental condition in the vehicle. For real-world applications, the head orientation will be estimated using automatic algorithms using either  RGB cameras \cite{Baltrusaitis_2018} or depth cameras \cite{Hu_2020}. 

\subsection{Calibration of Camera and Markers}
\label{ssec:Calibration}

A key challenge is to define a common coordinate system. We need the location and orientation of the driver's head along with the location of each enumerated marker in a 3D space with respect to a single coordinate system. Figure \ref{fig:ref_markers_a} shows the view from the rear camera facing the driver, and Figure \ref{fig:ref_markers_f} shows the view from the road camera. It is clear from these two figures that most of the markers are not included in the view of either of the camera. The problem is even more challenging as we aim to map the gaze direction to areas on the road camera. We need a calibration process to find the exact target marker location in the 3D space and the transformation between the cameras to represent all the coordinates in a single reference system. The calibration process relies on AprilTags to find the location of the markers in the 3D space and to find the relative homogeneous transformation between each camera. The proposed solution consists of placing AprilTags in the vehicle. The AprilTags are used to establish a connection between the road and face cameras, which do not have any overlap in their field of view. The calibration process has two steps: create a common reference coordinate system, and create a mapping between objects outside the vehicle.

The first step in the calibration is to establish a reference coordinate system. AprilTags are placed on each of the markers (Fig. \ref{fig:ref_markers_d}), and some reference locations in the field of view of the face camera (Figs. \ref{fig:ref_markers_a}-\ref{fig:ref_markers_d}). These tags are only used to calibrate the system, and are removed during the data collection. Then, we use a third camera to take multiple pictures containing subsets of these AprilTags (Fig. \ref{fig:ref_markers}). This camera captures locations that are not in the field of view of either of the dash cameras. The relationship between frames containing multiple common tags is calculated. The face camera captures a subset of these additional tags. Using the location of these tags, we create homogeneous transformations to obtain the location of all the tags, including the 21 markers, with respect to the coordinate system of the face camera.  

The second step in the calibration consists of estimating a mapping between the reference coordinate system and objects outside the vehicle. For this step, the third camera is fixed inside the vehicle such that it records the windshield and the road view (Figs. \ref{fig:ref_markers_e} and \ref{fig:ref_markers_f} -- these two images were simultaneously taken). As a result, we have two cameras facing the road: the road camera used in the data collection and the third camera used for calibration. We printed an AprilTag sign, which is placed in front of the vehicle such that it is in the field of view of both cameras. This process helps us to establish the relationship between the cameras for different points. Finally, the relation between the face camera and the third camera is established using the location of the markers. With this process, we derive all the transformations needed to map everything into the coordinate system of the face camera. The transformation matrix to relate each camera is calculated using the Kabsch algorithm \cite{kabsch_1978}.

Since the placement of the headband slightly varies across subjects, we need to obtain a standard reference per subject to estimate the head pose from the AprilTags. For this purpose, we assume that the long-term average of the driver's head pose is consistent across subjects. We calculate the average orientation of the headband in the quaternion space using \emph{spherical linear interpolation} (Slerp). We set the origin of the coordinate system as the average pose of the driver’s head by subtracting the average head position per participant and multiplying by the inverse of the average rotation matrix to normalize the orientation. We also subtract the average head position value from each of the target marker's location to apply a similar transformation to the target gaze. Finally, the ground truth gaze vector at a given instant is obtained by subtracting the target gaze location from the head position at the given instant, calculating the horizontal and vertical gaze angles from this vector.

\section{Methodology}
\label{sec:methodology}


This paper aims to create a probabilistic salient visual map describing the visual attention of the driver. We project this map onto the windshield creating spatial distributions for the gaze direction. Then, we map this visual map on the road camera, defining areas on the road where we predict the driver is directing his/her gaze. We can estimate confidence regions for the driver's visual attention by creating a probabilistic map, which is an appealing method with more practical applications than methods predicting a single point for the gaze direction. This section proposes our main method as well as three alternative baselines to obtain a probabilistic distribution of the gaze angle from the position and orientation of the driver's head. These methods estimate the probabilistic salient visual map for the horizontal and vertical angles by modeling the mean and variance of the gaze angles as a function of the position and orientation of the head.

\subsection{Gaussian Process Regression}
\label{ssec:gpr}

Our proposed model is based on the original design implemented in our preliminary work \cite{Jha_2017}. It relies on \emph{Gaussian process regression} (GPR) \cite{Rasmussen_2004}, which models the outputs as a Gaussian process with the co-variance defined by a kernel function. The model assumes that any subset of the output is a joint Gaussian distribution. Using the ground truth of the training data in the vicinity of each point, the model learns the uncertainty in the prediction of any test data. This method provides a promising and effective approach to learn the many-to-many relationship between the head pose and the gaze. It learns a gaze distribution as a function of different head poses presented in the training data that are in the neighborhood of the target head pose.

Let $\mathbf{x} \in \mathbbm{R}^{d}$ be the input of the system, where $d$ is its dimension (in our case $d=6$). Let $Y$ be a Gaussian random process representing the output. If $\mathbf{y}$ is a vector representing $n$ realizations, as a Gaussian random process, $\mathbf{y}$ follows a joint Gaussian distribution with prior distribution $f_{\mathbf{y}}$,

\begin{equation}
\label{eqn:1}
f_{\mathbf{y}} = \mathcal{N}(\mathbf{\mu},\Sigma)
\end{equation}

\noindent
where, $\mathbf{y} = \{y_{1},y_{2},y_{3},\ldots,y_{n}\} \subset Y$  ($y_{i}$ is a realization of $Y$). The parameters $\mathbf{\mu}$ and $\Sigma$ are functions of $\mathbf{x}$. As shown in Equation \ref{eqn:det}, the mean provides the deterministic component of the model, where $\mathbf{\omega} \in \mathbbm{R}^{d}$ and $\omega_{0}\in \mathbbm{R}$ are learned while training the models.

\begin{equation}
    \label{eqn:det}
    \mathbf{\mu} = \mathbf{x}^T\mathbf{\omega} + \omega_{0}
\end{equation}

The probabilistic component is given by $\Sigma$, which is the covariance matrix. The covariance of any point $\mathbf{x}$ calculated jointly with the input points in the training set $\mathbf{x'}$ is given by the kernel $k(\mathbf{x},\mathbf{x'})$. The covariance is modeled using a squared exponential kernel (Eq. \ref{eq:gpr}). The correlation is learned with respect to the input data from the training set in the neighborhood of the data of interest. This kernel imposes that the outputs will be more correlated to the points in the training data that are closer to the test input data as the covariance matrix will have higher values for points that are closer. 

\begin{equation}
k(\mathbf{x},\mathbf{x}') = \sigma^{2}_{f} \exp \left( \frac{-\|\mathbf{x}-\mathbf{x'}\|^{2}}{2l^{2}} \right)
\label{eq:gpr}
\end{equation}

In Equation \ref{eq:gpr}, the parameter $\sigma_{f}$ represents the amplitude of the covariance. This parameter defines the autocovariance of the data points (i.e., $k(\mathbf{x},\mathbf{x}) = \sigma_{f}^{2}$). The parameter $l$ represents the length scale value, which  defines how much the distance between the training and predicted data affects the cross-covariance between two data points. If $l$ is high, $k(\mathbf{x},\mathbf{x'})$ slowly reduces, as the distance between the points increases ($\|\mathbf{x}-\mathbf{x'}\|^{2}$). These parameters decide the size of the confidence interval of our prediction as the covariance matrix is a function of these parameters. 
We also explore the use of \emph{automatic relevance determination} (ARD). Using ARD, the kernel learns different length scale parameters for each input variable. The kernel function with ARD is given in Equation \ref{eq:gpr_ard}. 

\begin{equation}
k(\mathbf{x},\mathbf{x}') = \sigma^{2}_{f} \exp \left( -\frac{1}{2}\sum\limits_{i=0}^d \frac{\|x_i-x_{i}'\|^{2}}{l_{i}^{2}} \right)
\label{eq:gpr_ard}
\end{equation}

Using different values for $l$ may be useful, since the input to our models include position and orientation of the head, which may have different scales. We learn the values for $\sigma_{f}$ and $l$ (or $l_{i}$) while training the models by maximizing the log-likelihood of the ground truth data in the train set.

To obtain the posterior distribution from the prior model, the model is conditioned on the given training data. Let, $X_{tr} \in \mathbbm{R}^{L \times d}$ be the training dataset and $y_{tr} \in \mathbbm{R}^{L}$ be the output Gaussian random variables, where $L$ is the number of frames in the train set. Let $y_{*}$ be the random variable we are trying to estimate for the input vector $\mathbf{x}_{*}$. From Equation \ref{eqn:1}, the joint distribution is given by,

\fontsize{8.5}{10}\selectfont
\begin{align}
    \label{eqn:5}
    & \left[ \begin{array}{c} f_{y_{tr}} \\ f_{y_{*}} \end{array} \right] = \mathcal{N} \left( \left[ \begin{array}{c}
    X_{tr}^T\mathbf{\omega} + \omega_{0} \\
    \mathbf{x}_{*}^T\mathbf{\omega} + \omega_{0} \end{array} \right], \begin{bmatrix} \Sigma_{(X_{tr},X_{tr})} & \Sigma_{(X_{tr},\mathbf{x}_{*})} \\ \Sigma_{(\mathbf{x}_{*},X_{tr})} & \Sigma_{(\mathbf{x}_{*},\mathbf{x}_{*})} \end{bmatrix}\right)
\end{align}
\normalsize

Using equation \ref{eqn:5}, the posterior distribution can be calculated with the conditional probability when $y_{tr} = y_{obs}$:

\begin{equation}
\begin{aligned}
    f_{y_{*}|y_{tr}=y_{obs}} = \mathcal{N}(\hat{\mu}_{*},\hat{\Sigma}_{*}) \label{eqn:6}
\end{aligned}
\end{equation}
\begin{equation}
    \label{eqn:7}
    \hat{\mu}_{*} = \mathbf{x}_{*}^T\mathbf{\omega} + \omega_{0} + \Sigma_{(\mathbf{x}_{*},X_{tr})}[\Sigma_{(X_{tr},X_{tr})}]^{-1}(y_{obs}-X_{tr}^T\mathbf{\omega} - \omega_{0})
\end{equation}
\begin{equation}
    \label{eqn:8}
    \hat{\Sigma}_{*} = \Sigma_{(\mathbf{x}_{*},\mathbf{x}_{*})} -  \Sigma_{(\mathbf{x}_{*},X_{tr})}[\Sigma_{(X_{tr},X_{tr})}]^{-1}\Sigma_{(X_{tr},\mathbf{x}_{*})}
\end{equation}

\noindent 
where $y_{obs}$ is the ground truth value (i.e., observed $y$). We use four different settings for the deterministic function. The first setting is a GPR model without the deterministic component (i.e. $\mathbf{\omega}= 0, \omega_{0} = 0$). The mean of the posterior distribution is purely estimated from the kernel function (Eq. \ref{eqn:9}).

\begin{equation}
    \label{eqn:9}
    \hat{\mu}_{*} = \Sigma_{(\mathbf{x}_{*},X_{tr})}[\Sigma_{(X_{tr},X_{tr})}]^{-1}y_{obs}
\end{equation}

The second setting is with a constant deterministic component. The model learns $\omega_0$ as a single constant mean for the distribution (i.e., $\mathbf{\omega} = 0$). The third setting estimates both $\mathbf{\omega}$ and $\omega_{0}$ during training. We refer to this setting as linear model. The fourth setting estimates the deterministic component of the model with a \emph{neural network} (NN). We implement this approach by training  $\mathit{NN}(\mathbf{x})$, using back propagation. The network is implemented with two hidden layers, following the architecture used for our second baseline (Fig. \ref{fig:nn_model}). Then, we estimate the residual error, $r(\mathbf{x})=y_{obs}-\mathit{NN}(\mathbf{x})$, which is modeled with the GPR formulation, without the deterministic component ($\mathbf{\omega}$ = 0,  $\omega_{0}$ = 0). The  conditional mean for this implementation is given by Equation \ref{eqn:gpr_nn_u}. 

\begin{align}
    \label{eqn:gpr_nn_u}
    & \hat{\mu}_{*} = \mathit{NN}(\mathbf{x}_{*}) +\Sigma_{(\mathbf{x}_{*},X_{tr})}[\Sigma_{(X_{tr},X_{tr})}]^{-1}(y_{obs}-\mathit{NN}(X_{tr}))
\end{align}

Using this framework, we learn two separate models for the horizontal angle ($\theta$) and the vertical angle ($\phi$). An important feature of our formulation is modeling the output as a heteroscedastic process, where the variance of the output salient map varies depending on the input variables. Therefore, the size of the probabilistic salient visual map increases for regions with higher uncertainty, and decreases when the model is confident in its prediction.

\subsection{Baseline Methods}
\label{ssec:Baseline}

We compare the model with three methods. Two of these baselines are based on normal regression functions designed with the mean square error loss. We adapted these regression models to create a probability map as the output by assuming a Gaussian distribution. For the third baseline,  we explore a variation of \emph{mixture density network} (MDN) that uses the log-likelihood as the loss function to model the conditional probability density of the gaze given the input head pose. This section provides the details of these baseline models. 

\subsubsection{Linear Regression}
\label{sssec:reg1}

The first baseline is the most basic regression model. The gaze is obtained as a linear function of the head pose parameters (orientation and position). The dependent variables are the six degrees of freedom of the head corresponding to its position (x,y,z) and orientation angles ($\alpha$,$\beta$,$\gamma$). Two separate models are created for the gaze angle in the horizontal and vertical directions. Equations \ref{eq:linear_h} and \ref{eq:linear_v} show the models, where, $\theta_{gaze}$ and $\phi_{gaze}$ are the horizontal and vertical gaze angles, respectively.

\begin{equation}
\theta_{gaze} = a_0 + a_1x + a_2y + a_3z + a_4\alpha + a_5\beta + a_6\gamma
\label{eq:linear_h}
\end{equation}
\begin{equation}
\phi_{gaze} = b_0 + b_1x + b_2y + b_3z + b_4\alpha + b_5\beta + b_6\gamma
\label{eq:linear_v}
\end{equation}

This model is similar to the one trained in Jha and Busso \cite{Jha_2016}, but instead of obtaining the gaze location, we obtain the angles representing the gaze vectors. To create a probability distribution as our prediction, we consider a Gaussian distribution with the mean value provided by the regression models. The variance is obtained from the mean square error estimated on the train data. Notice that this model is homoscedastic, where the variance is constant across the data.

\subsubsection{Regression with neural network}
\label{sssec:reg2}

\begin{figure}[!t]
\centering
\subfigure[Neural Network]
{
\includegraphics[width=8cm]{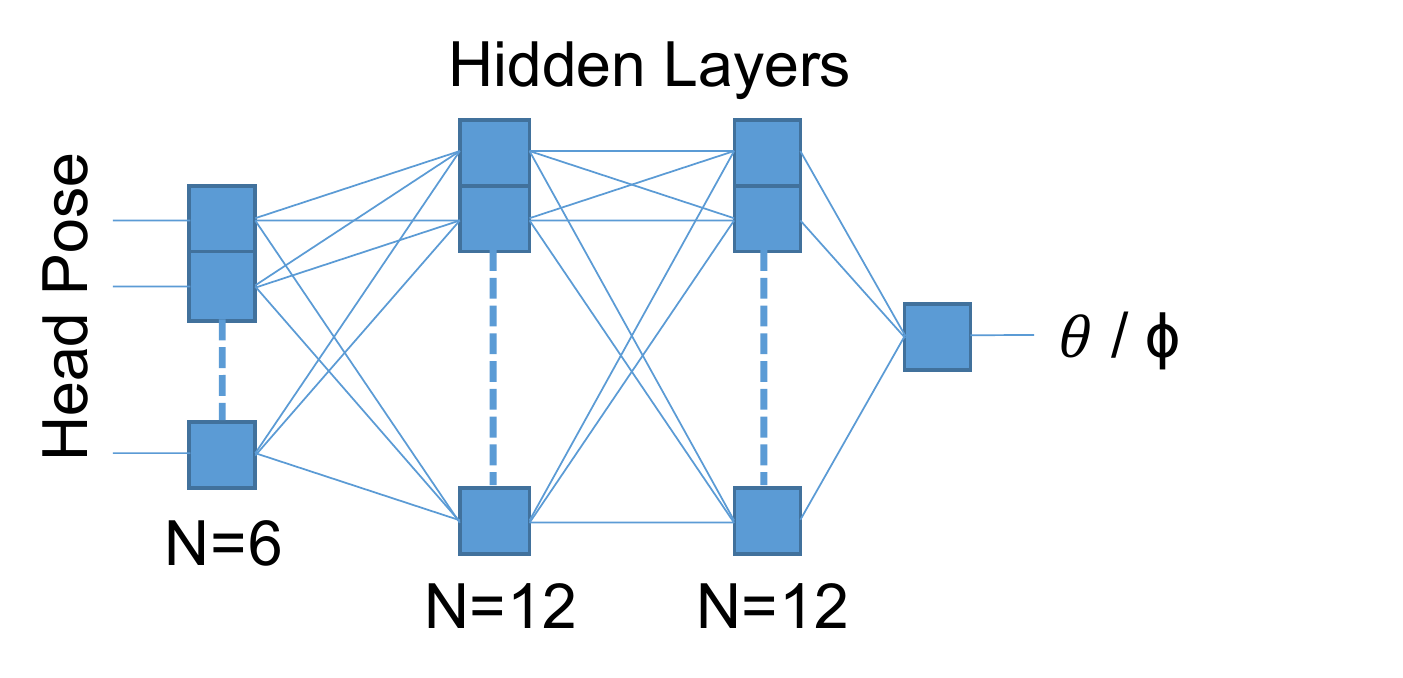}
\label{fig:nn_model}
}\\
\subfigure[Neural Network for Density Estimation]
{
\includegraphics[width=8cm]{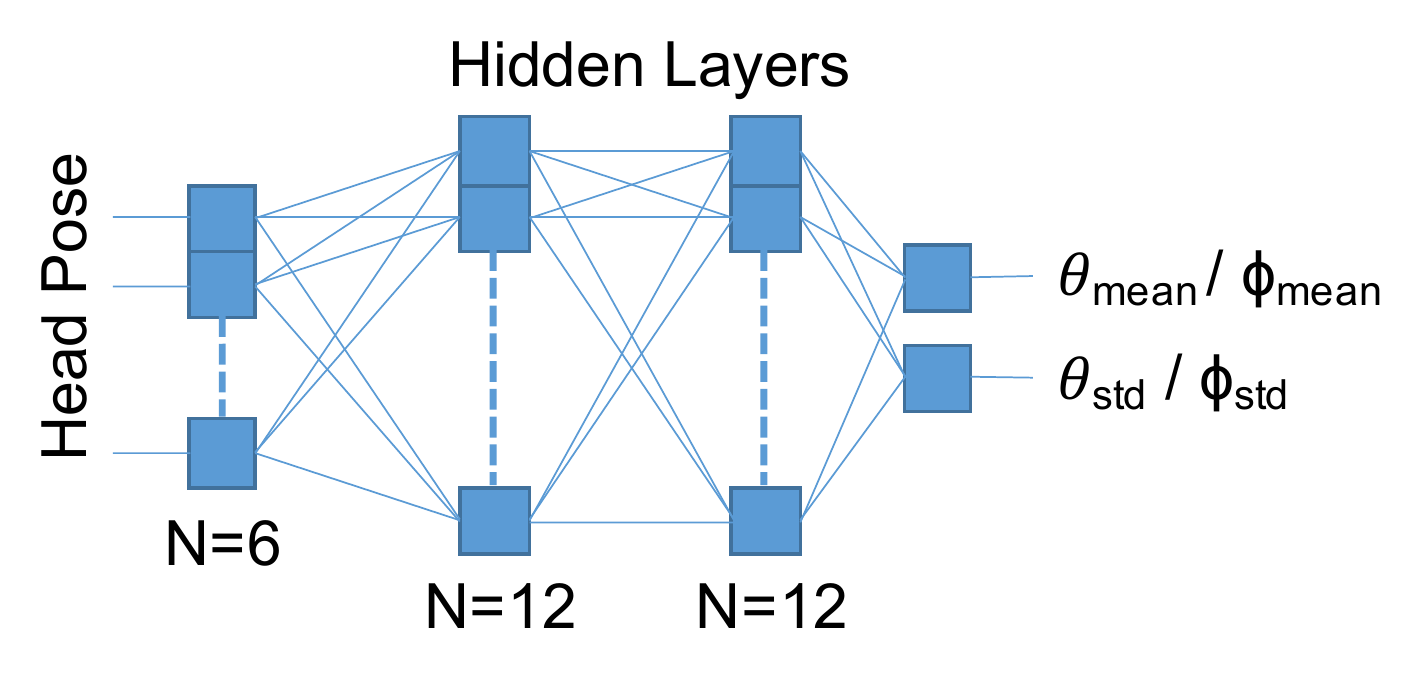}
\label{fig:mdn_model}
}
\caption{Architecture of baseline methods to estimate the probabilistic salient visual map describing visual attention. The same architecture is used for both horizontal and vertical angles.}
\label{fig:baseline_models}
\end{figure}

For the second baseline, we design a neural network to perform the regression task. Figure \ref{fig:nn_model} shows the model. The neural network contains two fully connected layers, each of them implemented with twelve nodes. The activation used for the hidden layers is the \emph{rectified linear unit} (ReLU) activation using a linear function at the output layer. The neural network is optimized to minimize the mean square error between the true gaze angle and the predicted gaze angle from the model. Similar to our previous baseline model, the probabilistic distribution is obtained by assuming a Gaussian distribution for the output, where the mean is the predicted gaze, and the variance is estimated with the mean square error in the train data. This approach is also homoscedastic.

\subsubsection{Neural Network for Density Estimation}
\label{sssec:reg4}
The third baseline is inspired by the MDN proposed by Bishop \cite{Bishop_1994}. MDN can directly learn the standard deviation of the output as a non-linear function of the input data. MDNs are used to model the output as a \emph{Gaussian mixture model} (GMM) by optimizing the log-likelihood function in Equation \ref{Eq:llk}.

\begin{equation}
p(y) = \Sigma^M_{k=1}\pi_k\mathcal{N}(y|\mu_k,\sigma_k)
\end{equation}

\begin{equation}
L_{llk}(y,\pi_k,\mu_k,\sigma_k) = -log(p(y))
\label{Eq:llk}
\end{equation}

The network has $3 \times M$ nodes in the output layer, where $M$ is the number of components. The output represents the component weights $\pi_k$, mean $\mu_k$ and standard deviation $\sigma_k$ with $k \in {1,\ldots,M}$. Since we assume that our output is a single Gaussian distribution, we design a model with one component, reducing the number of parameters to two. Therefore, the output layer has two nodes that provide the mean $\mu$ and the standard deviation $\sigma$. Our objective is to predict a Gaussian distribution that maximizes the probability of the ground truth data. To achieve this, we use as our loss function the negative log-likelihood of the ground truth gaze with respect to the predicted mean and the standard deviation.

\begin{equation}
L_{llk}(y,\mu,\sigma) =  -\log \left(  \frac{1}{\sqrt[]{2\pi}\sigma} \exp \left( \frac{(y - \mu)^2}{2\sigma^2}\right) \right)
\label{Eq:lossfuction}
\end{equation}

The variable $\sigma$ is obtained as the exponential of the corresponding output node to avoid the standard deviation from being negative. With this formulation, we estimate not only the mean, but also the variance in each prediction, providing an appropriate scaling to the uncertainty of each output prediction. This baseline is a heteroscedastic method, where the variance changes according to the input data.

Figure \ref{fig:mdn_model} shows the network architecture, which has two hidden layers implemented with 12 nodes. The network uses the Adam optimizer \cite{Kingma_2014_2} with a learning rate of  $r=0.001$, using mini batches of size 32. The neural network is  implemented in Keras \cite{Chollet_2015} with Tensorflow \cite{Abadi_2016} as backend. The networks is trained for 1,000 epochs, and the model with minimum validation loss is chosen as the final model to be evaluated in the test set.

\section{Experimental Evaluation}
\label{sec:Results}

This section evaluates the proposed solution and baselines to estimate the probabilistic salient visual map. The models are separately trained and evaluated for data collected in phase 1 (parked vehicle) and phase 2 (driving condition). The database is partitioned into train, test and validation set using a leave-one-driver-out cross-validation approach. Data from one subject are used for the validation set, data from one subject are used for the test set, and data from the remaining fourteen subjects are used for the training set. This approach is repeated sixteen times, where we report the results across the 16 folds. Note that all the data are, at some point, part of the test set.

We need to analyze the predicted probabilistic salient visual map in terms of accuracy and spatial resolution to evaluate and compare the effectiveness of the baseline and proposed models. 

Accuracy is measured as the percentage of the target gaze directions included in a given confidence interval. If the majority of the data do not lie within the confidence interval, the model is not accurate. The spatial resolution determines how large the confidence interval is.  Likewise, if the spatial resolution is too high, the prediction is not very useful even if most data lies within the interval. To evaluate the spatial resolution of the system, we evaluate the size of the confidence interval created by each model. The outputs of the model are horizontal ($\theta$) and vertical ($\phi$) angles. Therefore, we express the area of the confidence region in terms of the fraction of a sphere surrounding the driver's head. An ideal approach will create a confidence interval that is both accurate and with reduced spatial resolution. To analyze the tradeoff between accuracy and spatial resolution, we present plots with the accuracy of our model at different spatial resolution (Figs. \ref {fig:res_gpr}, \ref{fig:res_baseline}, \ref{fig:lim_gpr}). 

The first evaluation considers different implementations of the GPR model with different parameters to establish the best method for our purpose (Sec. \ref{ssec:gpr_tuned}). Then, we compare the best performing GPR models with the three alternative baselines (Sec. \ref{ssec:bl}). Then, we demonstrate the features of the model by projecting the confidence regions onto the windshield (Sec. \ref{sec:ResultsWindshield}), and road camera (Sec. \ref{ssec:ResultsRoad}). Finally, we study the performance when we have the orientation of the driver's head, but limited information about the head's position, which is a possible scenarios if regular cameras are used to estimate the head information (Sec \ref{ssec:ResultsLimited}).

\subsection{GPR Model Selection}
\label{ssec:gpr_tuned}

\begin{figure}[!t]
\centering
    \includegraphics[width=0.98\columnwidth,trim = 16cm 20.5cm 14cm 0cm, clip]{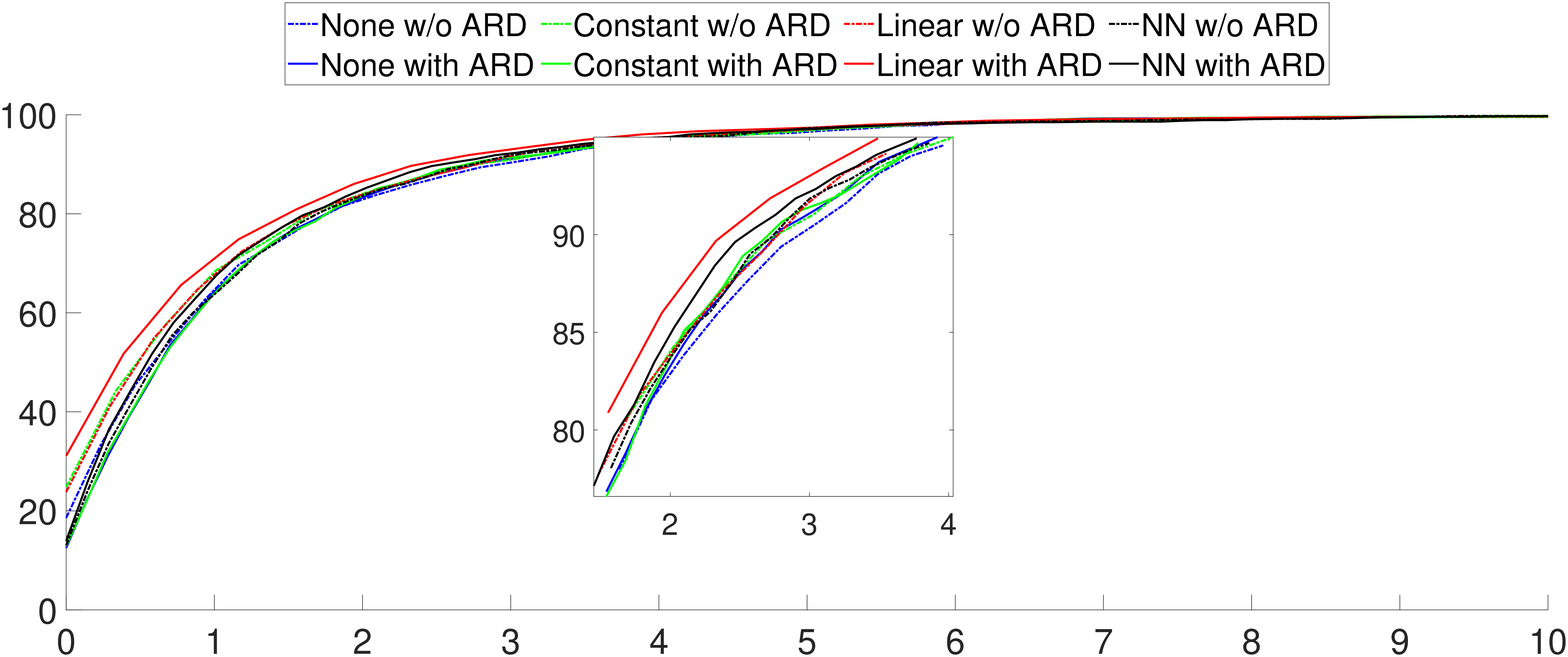}
\subfigure[Parked conditions]{
    \includegraphics[width=0.98\columnwidth,trim = 5.0cm 0cm 5.5cm 0cm, clip]{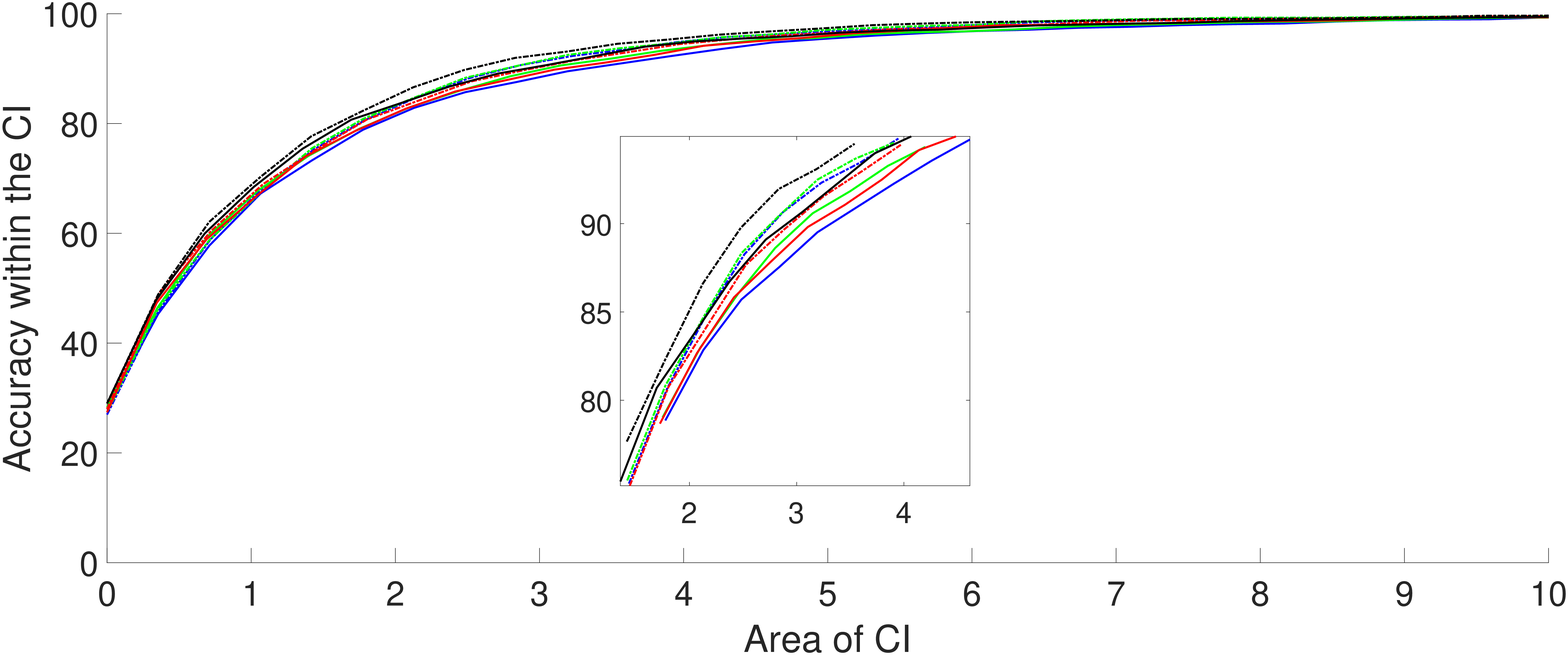}
    \label{fig:res_gpr_a}
}
\subfigure[Driving Conditions]{
    \includegraphics[width=0.98\columnwidth,trim = 5.0cm 0cm 5.5cm 0cm, clip]{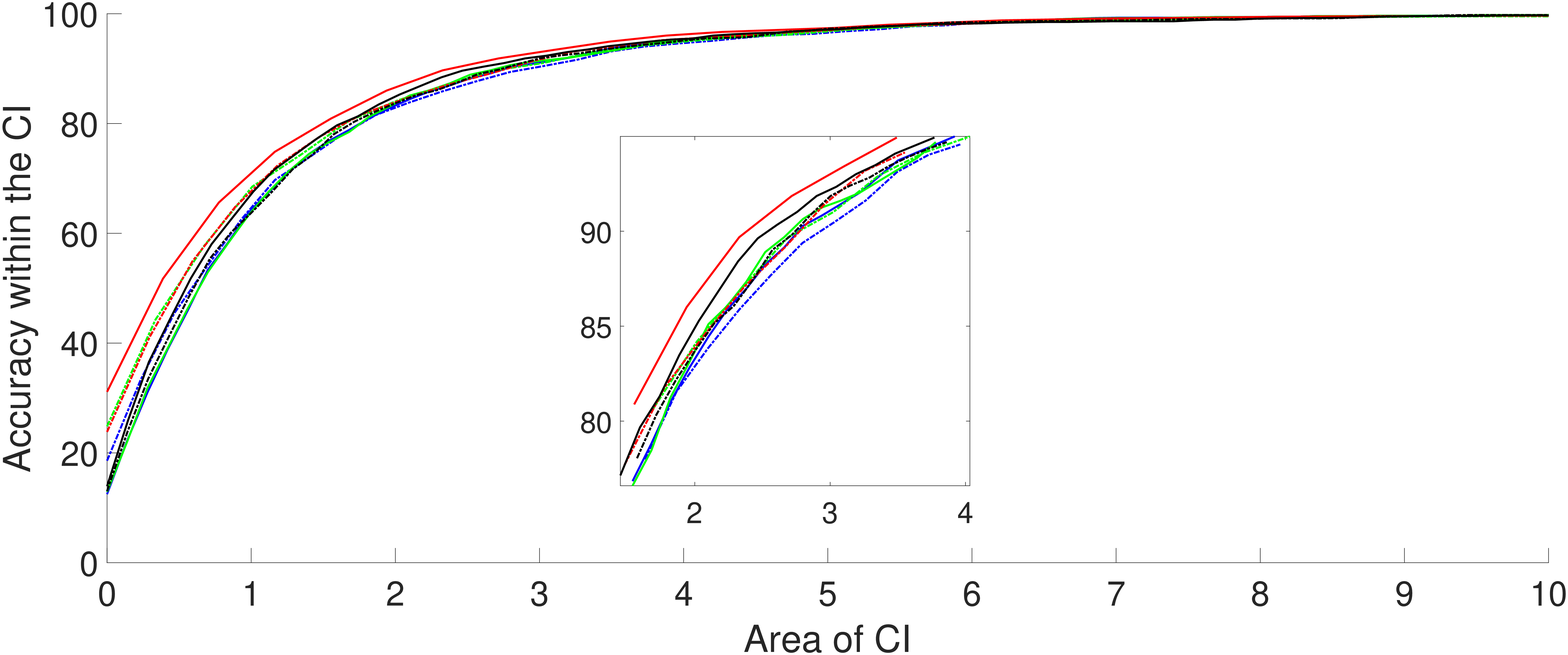}
    \label{fig:res_gpr_b}
}
\caption{Comparison of the accuracy versus temporal resolution of different implementations of the GPR model. We zoom the plots for better visualization.  The results are separately reported for parked and driving conditions. The figure is better viewed in colors.}
\label{fig:res_gpr}
\end{figure}

Figure \ref{fig:res_gpr} shows the accuracy of our model within different confidence interval for the different GPR models. Figure \ref{fig:res_gpr_a} reports the results for phase 1 (parked condition) and Figure \ref{fig:res_gpr_b} reports the results for phase 2 (driving condition). We zoom these figures between the 75\% and 95\% confidence intervals for better visualization. We observe that different models work better for parked and driving conditions. We have shown that the relationship between head movements and gaze changes when a person is driving \cite{Jha_2016}, which explain the difference in patterns across phases 1 and 2. During the parked condition, the best GPR model is when the deterministic part is implemented with a neural network, and the kernel is implemented without ARD. We consistently observe lower performance when using ARD, regardless of the implementation of the deterministic function. During the driving condition, implementing the deterministic component with a linear model leads to the best performance. For this case, the use of ARD in the kernel function leads to improvements for all four conditions. Since the relationship between head pose and gaze is more ambiguous when driving, as noted in Jha and Busso \cite{Jha_2016}, it is beneficial to add more flexibility to the model in phase 2. For phase 1, adding a more powerful deterministic function is enough to achieve good performance.

For the rest of the evaluation, we will consider the two GPR models that led to the best performance for phase 1 (GPR with neural network model without ARD) and phase 2 (GPR with linear model with ARD).

\subsection{Comparison with Baselines}
\label{ssec:bl}

This section compares our proposed models with the three baselines described in section \ref{ssec:Baseline}: \emph{linear regression} (LR), \emph{neural network regression} (NN) and \emph{mixture density network} (MDN). 

\subsubsection{Accuracy versus Spatial Resolution}
\label{sssec:bl_acc}

\begin{figure}[!t]
\centering
\subfigure[Parked conditions]{
    \includegraphics[width=0.98\columnwidth,trim = 3.5cm 0cm 5.5cm 0cm, clip]{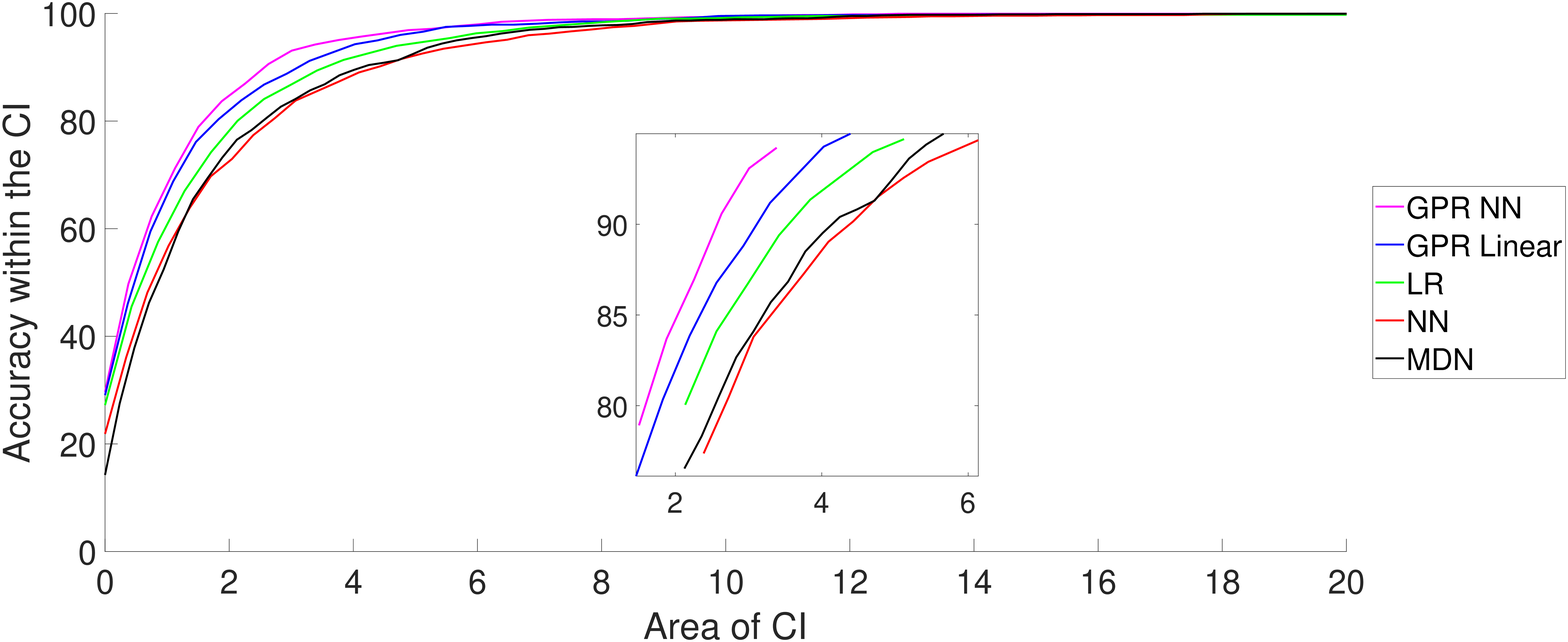}
    \label{fig:bsprk}
}
\subfigure[Driving Conditions]{
    \includegraphics[width=0.98\columnwidth,trim = 3.5cm 0cm 5.5cm 0cm, clip]{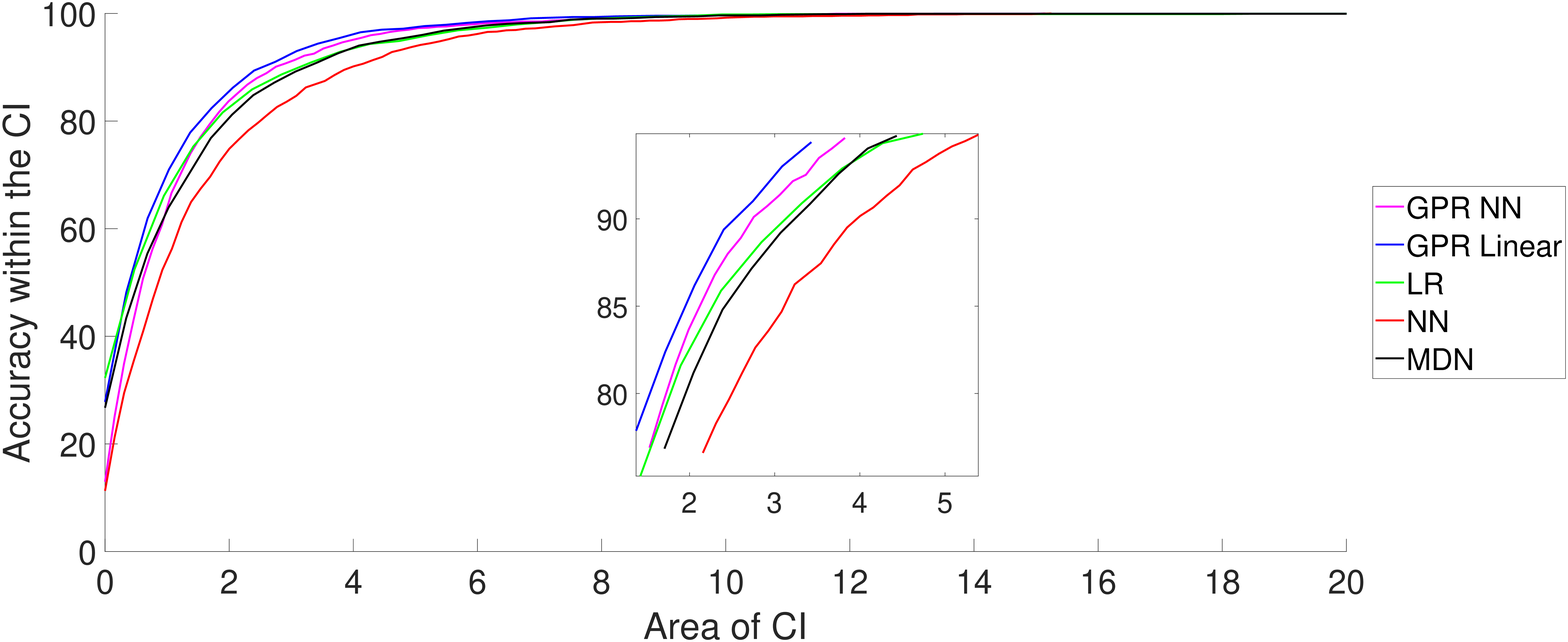}
    \label{fig:bsdrv}
}
\caption{Comparison of the accuracy versus temporal resolution of GPR and baseline models. We zoom the plots for better visualization.  The results are separately reported for parked and driving conditions. The figure is better viewed in colors.}
\label{fig:res_baseline}
\end{figure}

Figure \ref{fig:res_baseline} shows the accuracy of our models at different spatial resolutions, comparing with results with the curves of different baseline models. We observe that both GPR models perform better than all the baseline models. They are consistently above other curves showing not only higher accuracies, but also smaller regions. The linear regression baseline is the model with higher performance from the baselines. The values are constantly below our two implementations of the GPR models. 

To quantify the spatial resolution of the models, Table \ref{table:area} lists the area of the confidence interval at $50\%$, $75\%$ and $95\%$ accuracies for the baseline and GPR models. We observe that the areas of the confidence interval for the GPR models are smaller than the areas for the baseline models. In phase 1, GPR NN has the smallest area for the 95\% confidence interval (3.76\%). In phase 2, GPR Linear has the smallest area for the 95\% confidence interval (3.77\%). The ability to provide high accuracy within a small region makes the GPR models more efficient.

\begin{table}[t]
\caption{Average area of the confidence intervals for 50\%, 75\% and 95\% accuracy. This area is measured as the fraction of a sphere surrounding the driver's head.}
\begin{center}
\begin{tabular*}{0.99\columnwidth}{@{\extracolsep{\fill}}l|c|c|c|c|c|c}
\hline
\multirow{3}{*}{Method} & \multicolumn{3}{c}{Parked -- Phase 1} \vline & \multicolumn{3}{c}{Driving -- Phase 2}  \\ \cline{2-7}
                                   & {50\%} & {75\%} & {95\%} & {50\%} & {75\%} & {95\%}\\
                                  & [\%] & [\%] & [\%] & [\%] & [\%] & [\%] \\
\hline
LR &0.43& 1.71& 5.12& 0.47& 1.42& 4.74 \\ 
NN &0.68& 2.05& 6.48& 0.92& 2.00& 5.39\\ 
MDN &0.94& 2.12& 5.66& 0.68& 1.71& 4.43\\ 
\hline
GPR NN &0.38& 1.13& 3.76& 0.61& 1.38& 3.98\\
GPR Linear &0.37& 1.46& 4.39& 0.34& 1.37& 3.77\\ 
\hline
\end{tabular*}
\end{center}
\label{table:area}
\end{table}

To quantify the accuracy of the models, Table \ref{table:accuracy} lists the accuracy observed when the fractions of a sphere surrounding the driver's head is $1\%$, $2\%$ and $4\%$. This analysis quantifies the performance of the proposed and baseline models when their confidence intervals have consistent area. We observe that we can get $86.2\%$ accuracy in phase 2 within an area of $2\%$ with the GPR Linear model. Similarly on phase 1, we can obtain an accuracy of $83.7\%$ with the GPR NN model.

\begin{table}[t]
\caption{Average accuracy of the confidence intervals for probabilistic salient visual maps of different sizes (1\%, 2\% and 4\% of the sphere surrounding the driver's head).}
\begin{center}
\begin{tabular*}{0.99\columnwidth}{@{\extracolsep{\fill}}l|c|c|c|c|c|c}
\hline
\multirow{3}{*}{Method} & \multicolumn{3}{c}{Parked -- Phase 1} \vline & \multicolumn{3}{c}{Driving -- Phase 2}  \\ \cline{2-7}
                                   & {1\%} & {2\%} & {4\%} & {1\%} & {2\%} & {4\%}\\
                                  & [\%] & [\%] & [\%] & [\%] & [\%] & [\%] \\
\hline
LR &57.5& 80.1& 91.4& 66.1& 81.6& 92.9 \\ 
NN &56.8& 73.0& 89.0& 52.4& 74.8& 90.2\\ 
MDN &52.4& 73.2& 89.5& 63.9& 81.2& 94.0\\ 
\hline
GPR NN &71.2& 83.7& 95.7& 66.7& 83.7& 95.1\\
GPR Linear &68.8& 80.4& 94.3& 71.0& 86.2& 96.5\\ 
\hline
\end{tabular*}
\end{center}
\label{table:accuracy}
\end{table}

\subsubsection{Theoretical versus Empirical Cumulative Density Function}
\label{sssec:bl_cdf}

\begin{figure}[!t]
\centering
\subfigure[Parked conditions]{
    \includegraphics[width=0.98\columnwidth,trim = 21cm 0cm 16cm 0cm, clip]{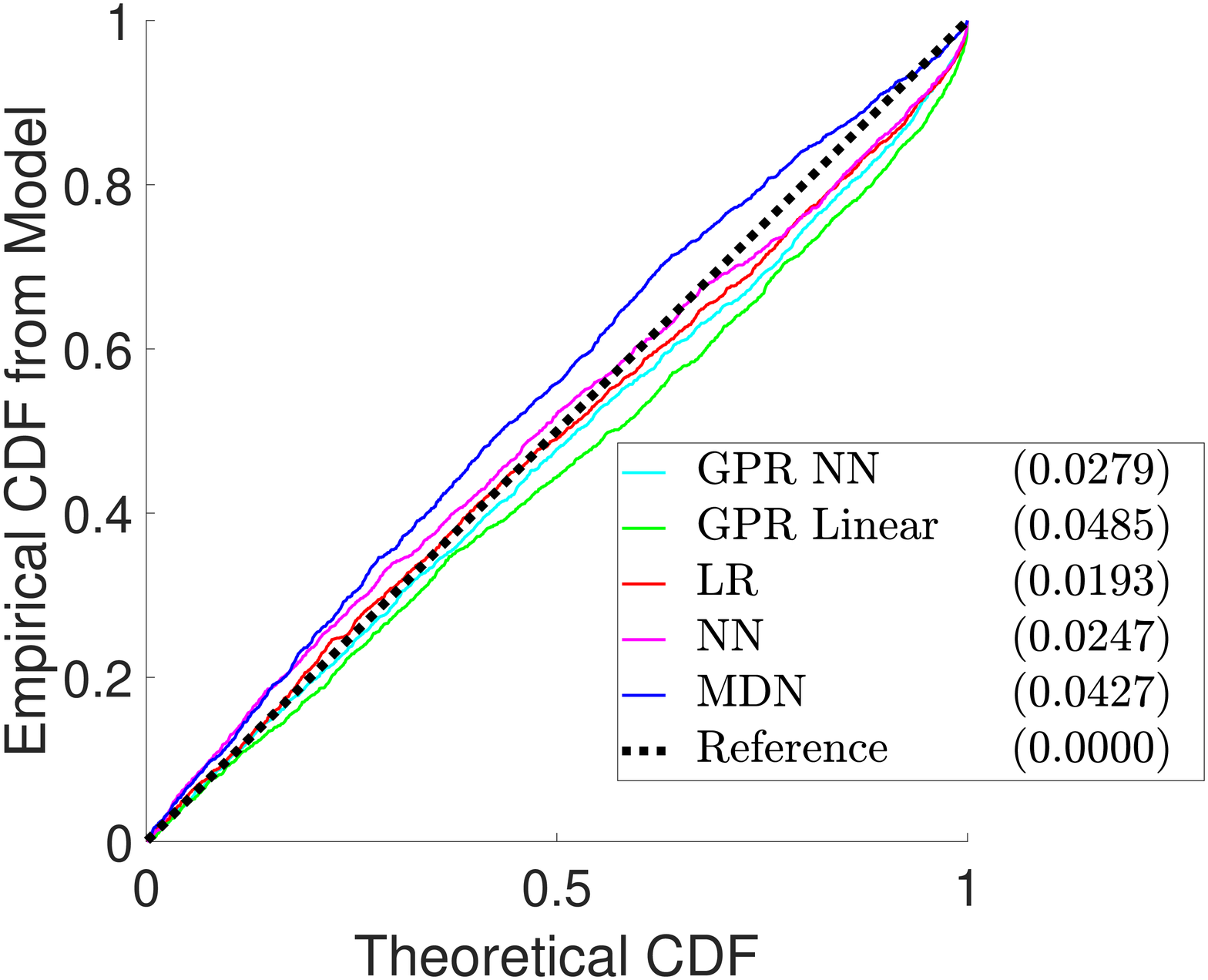}
    \label{fig:res_cdf_a}
}
\subfigure[Driving Conditions]{
    \includegraphics[width=0.98\columnwidth,trim = 21cm 0cm 16cm 0cm, clip]{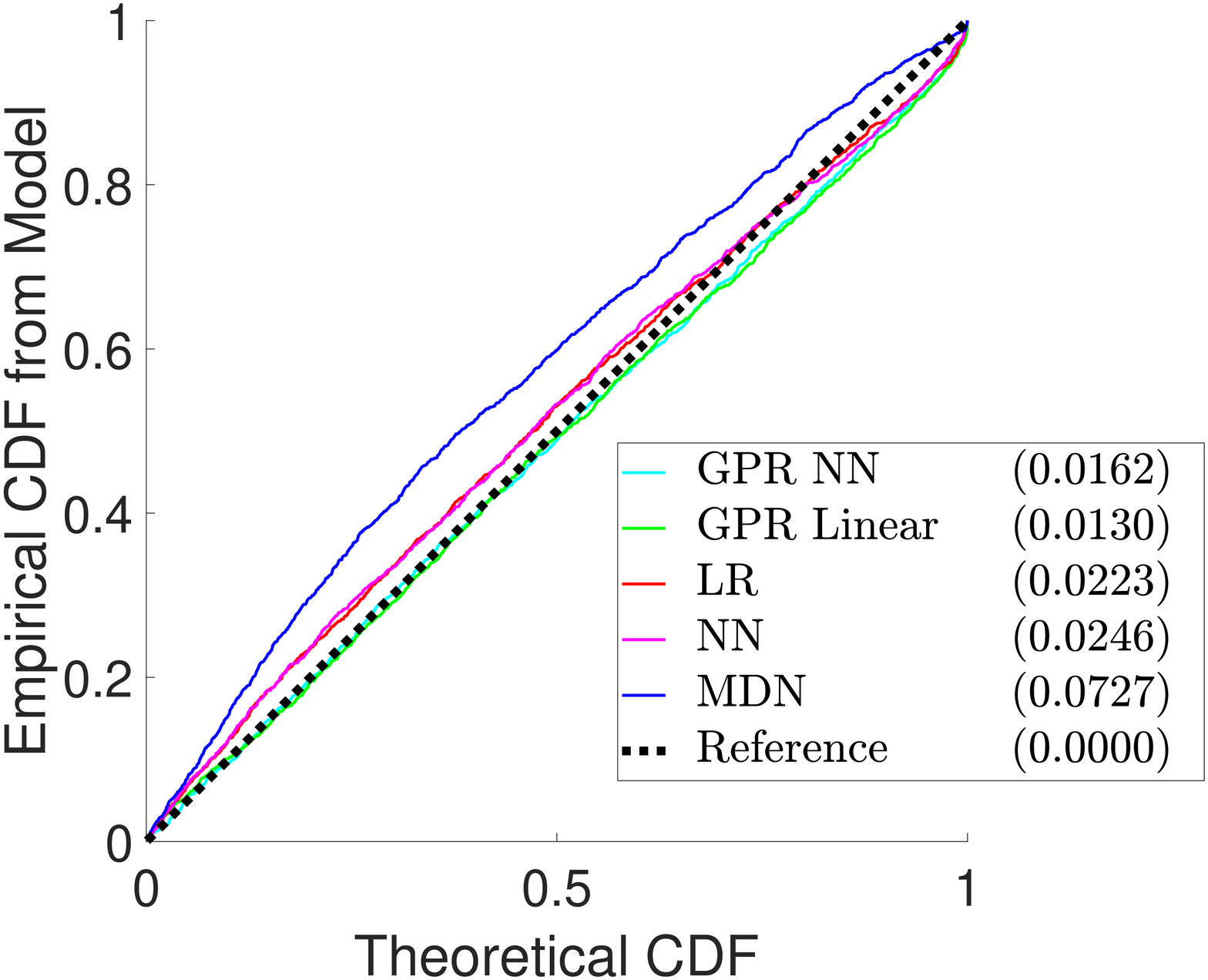}
    \label{fig:res_cdf_b}
}
\caption{Theoretical versus empirical cumulative distribution function for the GPR and baseline models. This figure evaluates whether the resulting probabilistic salient visual maps cover the target gaze direction as predicted by the Gaussian assumption in the models. The numbers in the legend quantify the fit using Equation \ref{Eq:cdf}.}
\label{fig:res_cdf}
\end{figure}

Since our proposed and baseline models assume that the predicted gaze follow a Gaussian distribution, it is important to analyze how well this Gaussian assumption holds with respect to the empirical distribution of the ground truth data around the predictions. For this analysis, we plot the fraction of the data observed within the confidence region (y-axis) as a function of the theoretical \emph{cumulative density function} (CDF) of the region (x-axis). Figure \ref{fig:res_cdf} shows the results for the parked and driving conditions. Ideally, we should observe the curves as close as possible as the reference diagonal curve (black curve). We measure the absolute area between each curve and the reference diagonal curve using Equation \ref{Eq:cdf}. The legend in Figure \ref{fig:res_cdf} reports the results.

\begin{equation}
\mathit{area} =  \frac{1}{N}\sum |\mathit{cdf}_{\mathit{theoretical}} - \mathit{cdf}_{\mathit{empirical}}|
\label{Eq:cdf}
\end{equation}

In the parked condition, the LR model is the closest to the reference diagonal curve. Since the distribution of the data is structured, a simple linear regression model with constant mean square error is enough to properly match the theoretical distribution for the confident intervals, although with lower accuracies and spatial resolutions than our proposed models (Tables \ref{table:area} and \ref{table:accuracy}). In the driving condition, the two GPR models are very close to the theoretical curve. The absolute areas from the reference diagonal curve are smaller than the corresponding absolute area for baseline models. Therefore, the GPR models not only provide better tradeoff for accuracy and spatial resolution, but also offer confidence intervals that are closer to the theoretical confidence intervals for the most important condition (phase 2). 

\subsection{Mapping the Confidence Regions onto the Windshield}
\label{sec:ResultsWindshield}

\begin{figure}[!t]
\centering
\subfigure[Face camera]
{
\includegraphics[height=2.1cm,trim = 15cm 7cm 11cm 0cm, clip]{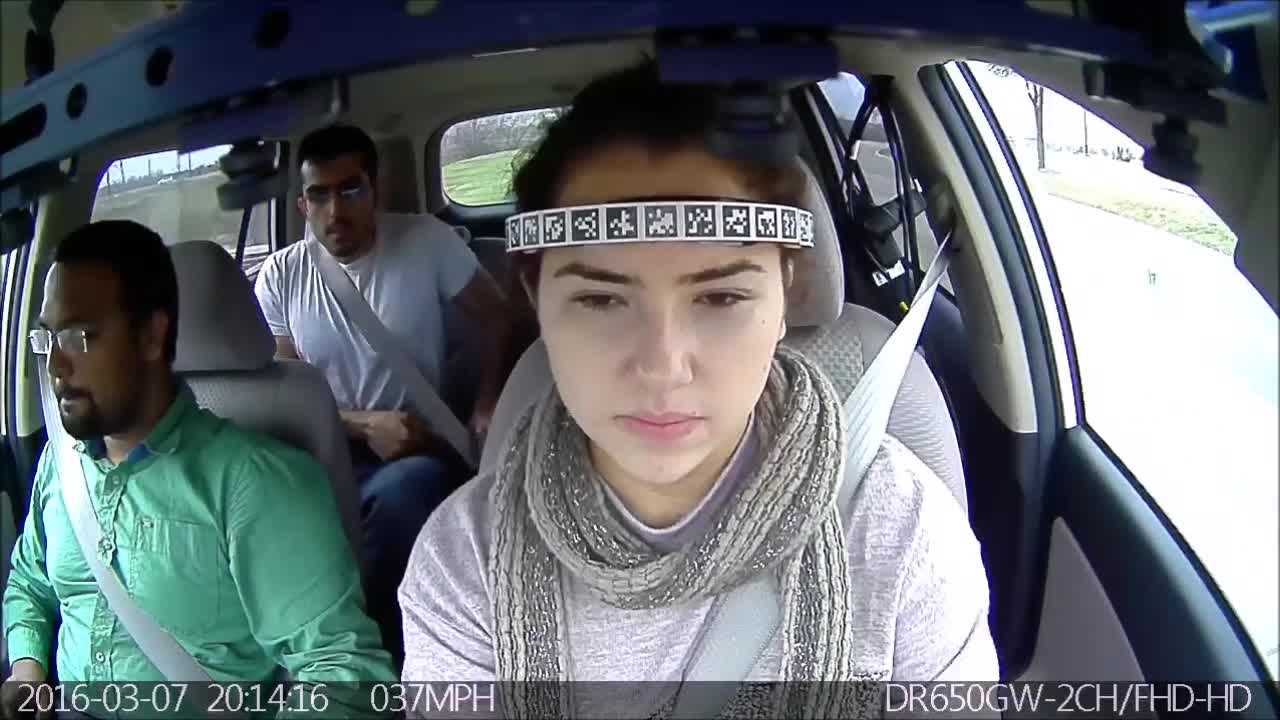}
\label{fig:mappingmarker_f1}
}\hspace{-0.3cm}
\subfigure[Windshield -- GPR]
{
\includegraphics[height=2.1cm,trim = 20cm 9cm 16cm 6cm, clip]{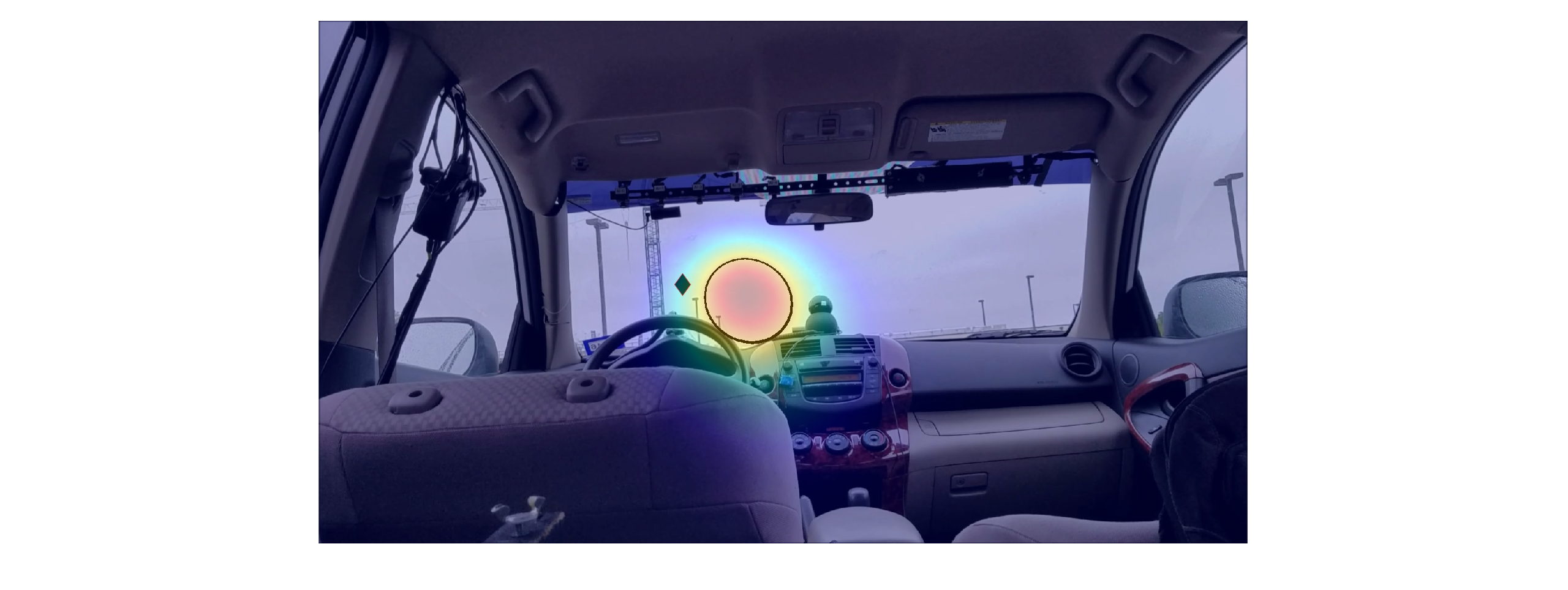}
\label{fig:mappingmarker_g1}
}\\
\subfigure[Face camera]
{
\includegraphics[height=2.1cm,trim = 15cm 7cm 11cm 0cm, clip]{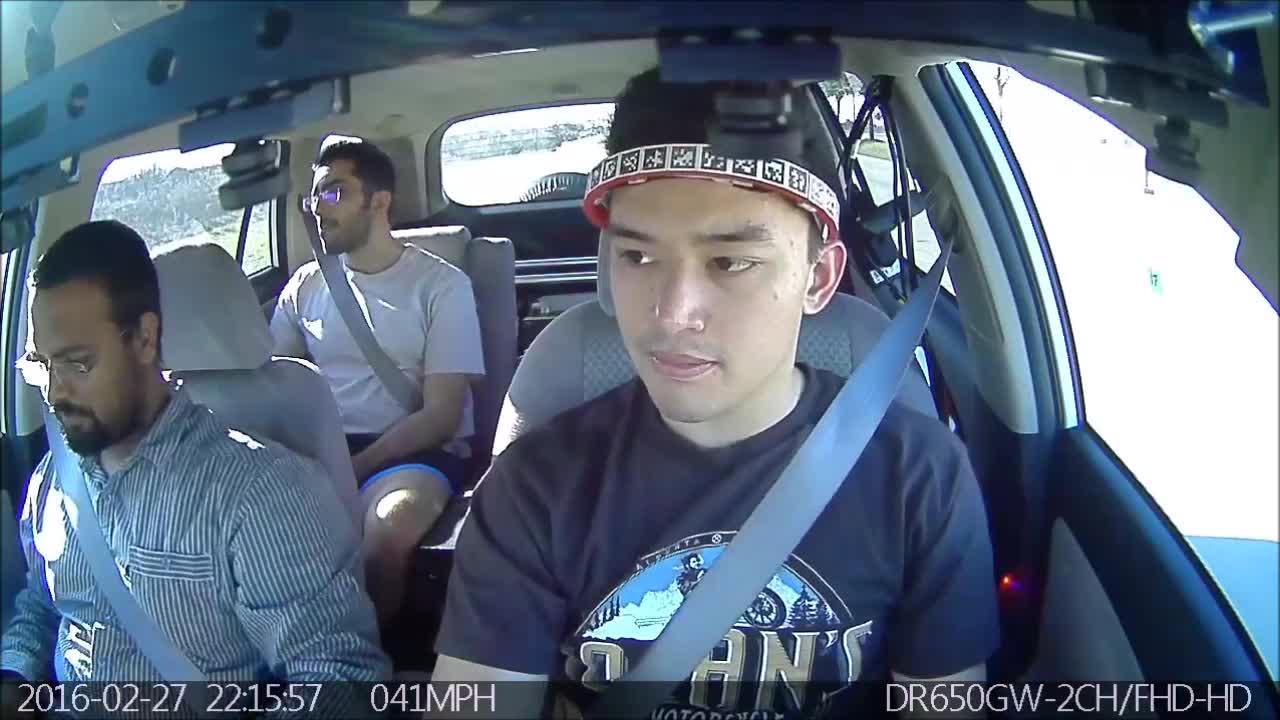}
\label{fig:mappingmarker_f2}
}\hspace{-0.3cm}
\subfigure[Windshield -- GPR]
{
\includegraphics[height=2.1cm,trim = 20cm 9cm 16cm 6cm, clip]{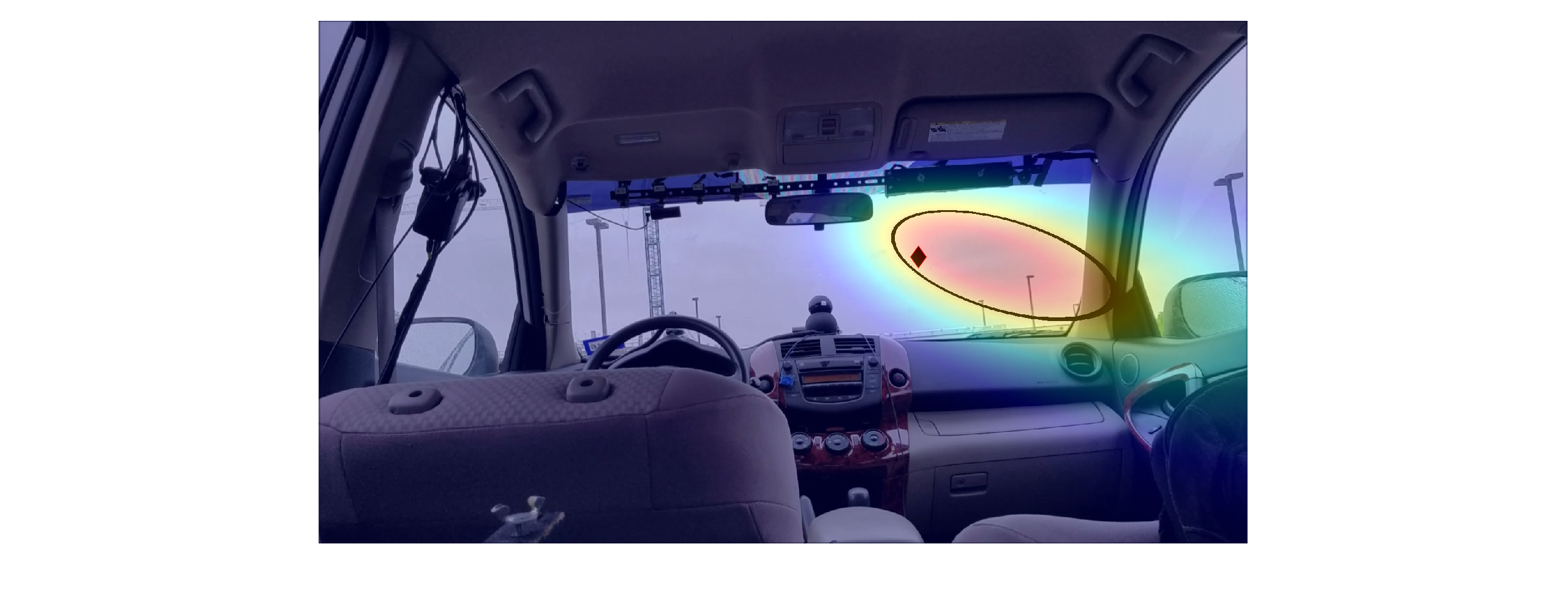}
\label{fig:mappingmarker_g2}
}\\
\caption{Two examples of projections of the probabilistic salient visual maps into the windshield. The target marker is highlighted with a black diamond. The darkened curves represent 50\%  confidence intervals.}
\label{fig:mappingmarker}
\end{figure}

This section projects the predicted confidence regions onto the windshield. We have the marker position, which is used as the ground truth for the gaze. We only use the targets markers from \#1 to \#13 for this purpose (Fig. \ref{Fig:markers}). We model the windshield as a plane by fitting the best plane containing these thirteen points. Small errors are introduced because the windshield is slightly curved so the points do not exactly lie on a plane. Therefore, when we project the original points back to the camera, they do not exactly match the target marker location (Fig. \ref{fig:mappingmarker}). From the gaze angles ($\alpha$ and $\beta$), the gaze direction is obtained by estimating the line from the position of the head ([$x_{hp}$,$y_{hp}$,$z_{hp}$]) towards the direction provided by the gaze vector. Equation \ref{eq:gaze_line} provides the projection used in the study. We estimate the region where a line meets the windshield plane. The probability density function at each point is calculated based on the probabilistic salient visual map created by the models.

\begin{multline}
[x,y,z] =[x_{hp},y_{hp},z_{hp}]\\
+ [\sin(\alpha), \cos(\alpha)\sin(\beta), \cos(\alpha)\cos(\beta)]
\label{eq:gaze_line}
\end{multline}

Figure \ref{fig:mappingmarker} shows two examples for the confidence regions created with the GPR model. While these are just two examples, they are representative of the probabilistic salient visual map created by the models. These figures also demonstrate how the predicted angles can be mapped onto the real world coordinates. The figure shows that the confidence regions in front of the drivers are smaller than the confidence regions on the side of the windshield, signaling more uncertainty. The size of the regions is learned from the data.

\begin{figure}[!t]
\centering
\subfigure[Face camera]
{
\includegraphics[height=2.4cm,trim = 15cm 5cm 11cm 2cm, clip]{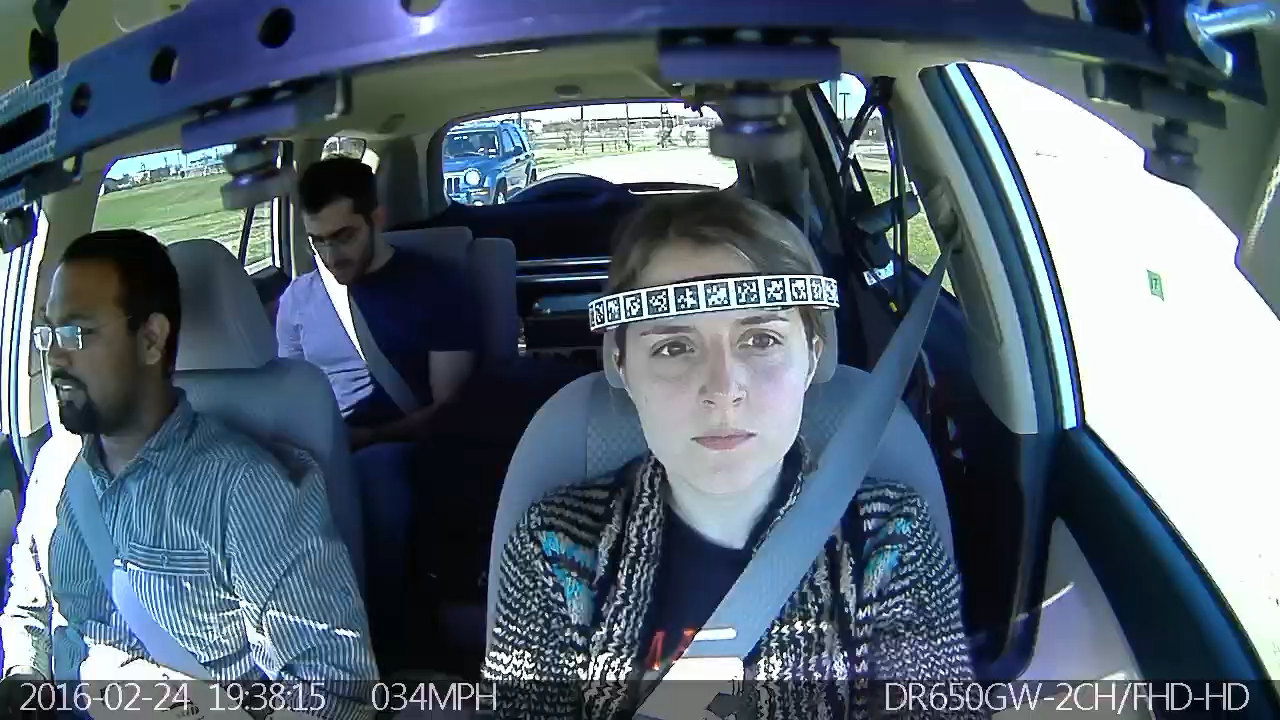}
\label{fig:mappingroad_f1}
}\hspace{-0.3cm}
\subfigure[Mapping on road, GPR]
{
\includegraphics[height=2.4cm,trim = 2cm 2cm 2cm 2cm, clip]{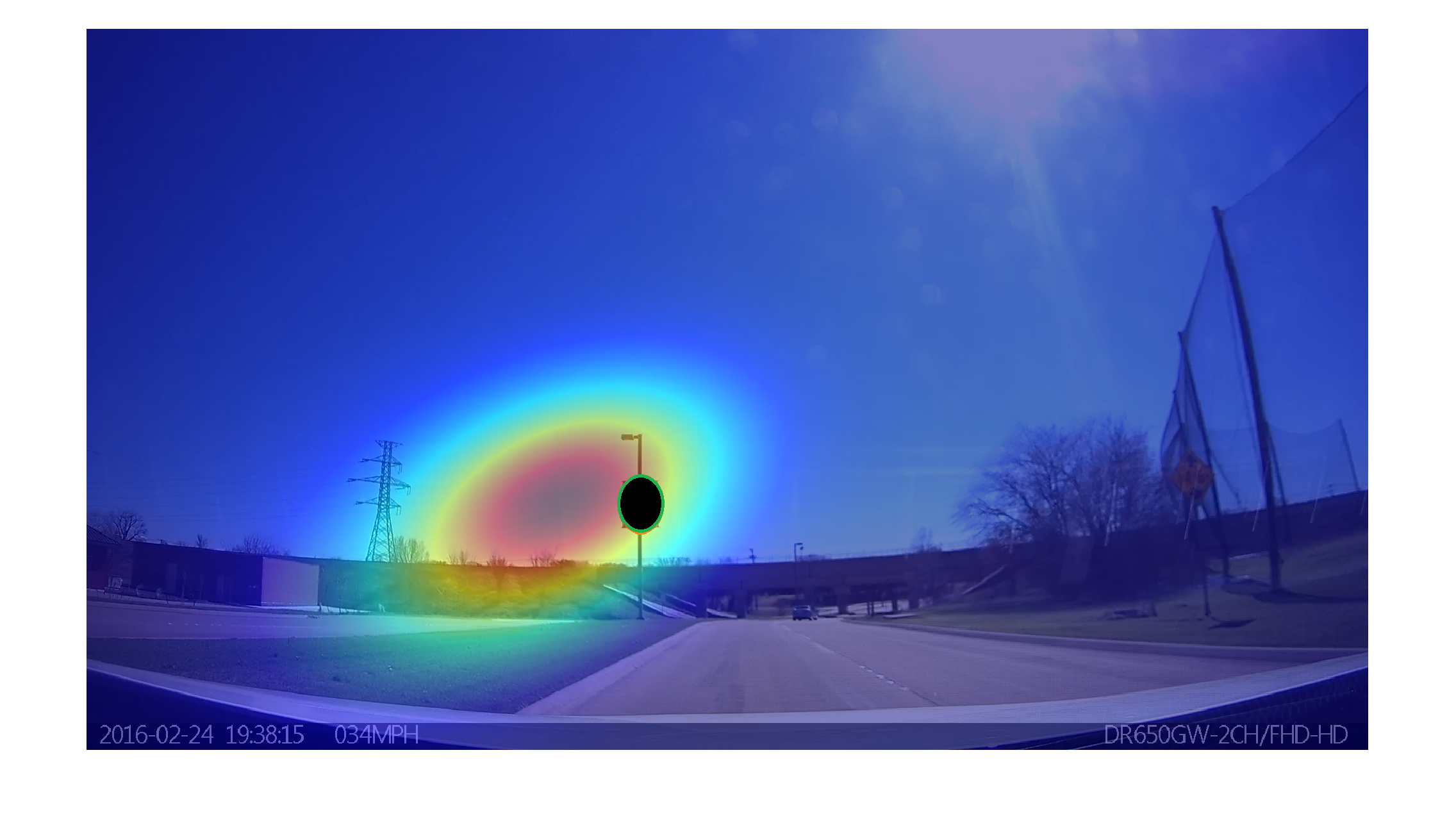}
\label{fig:mappingroad_g1}
}\\
\subfigure[Face camera]
{
\includegraphics[height=2.4cm,trim = 15cm 6cm 11cm 1cm, clip]{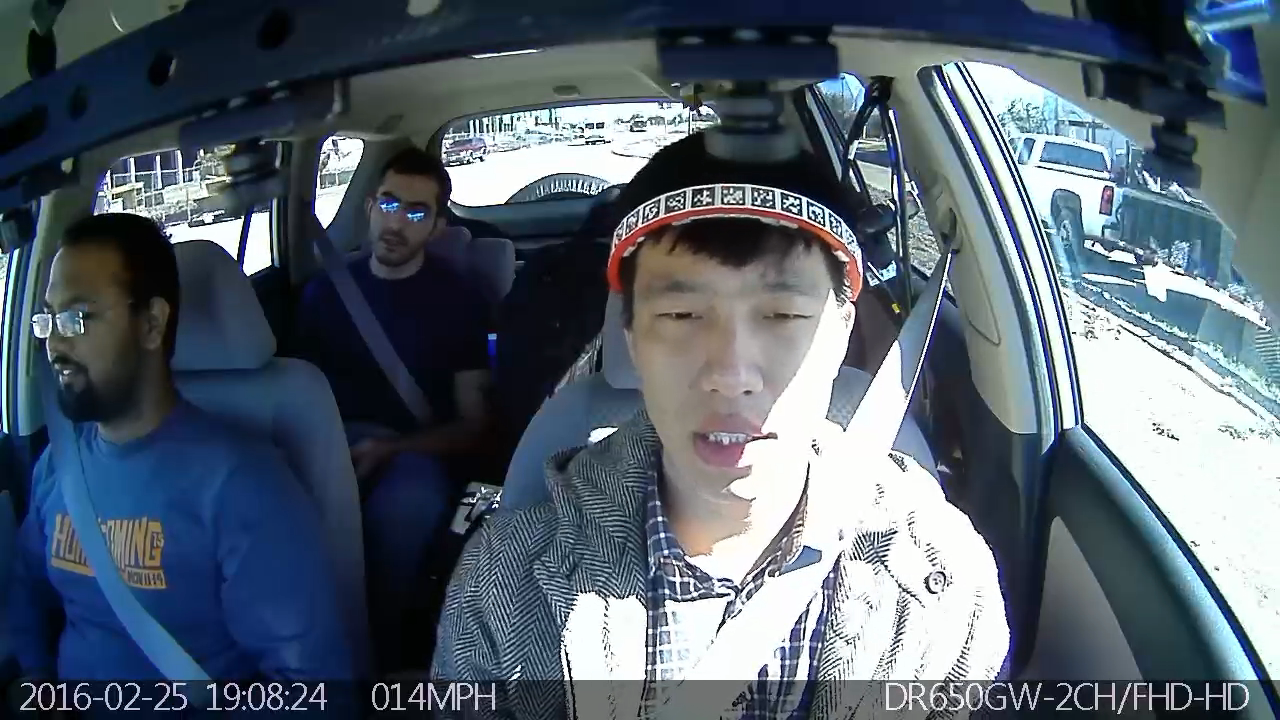}
\label{fig:mappingroad_f2}
}\hspace{-0.3cm}
\subfigure[Mapping on road, GPR]
{
\includegraphics[height=2.4cm,trim = 2cm 2cm 2cm 2cm, clip]{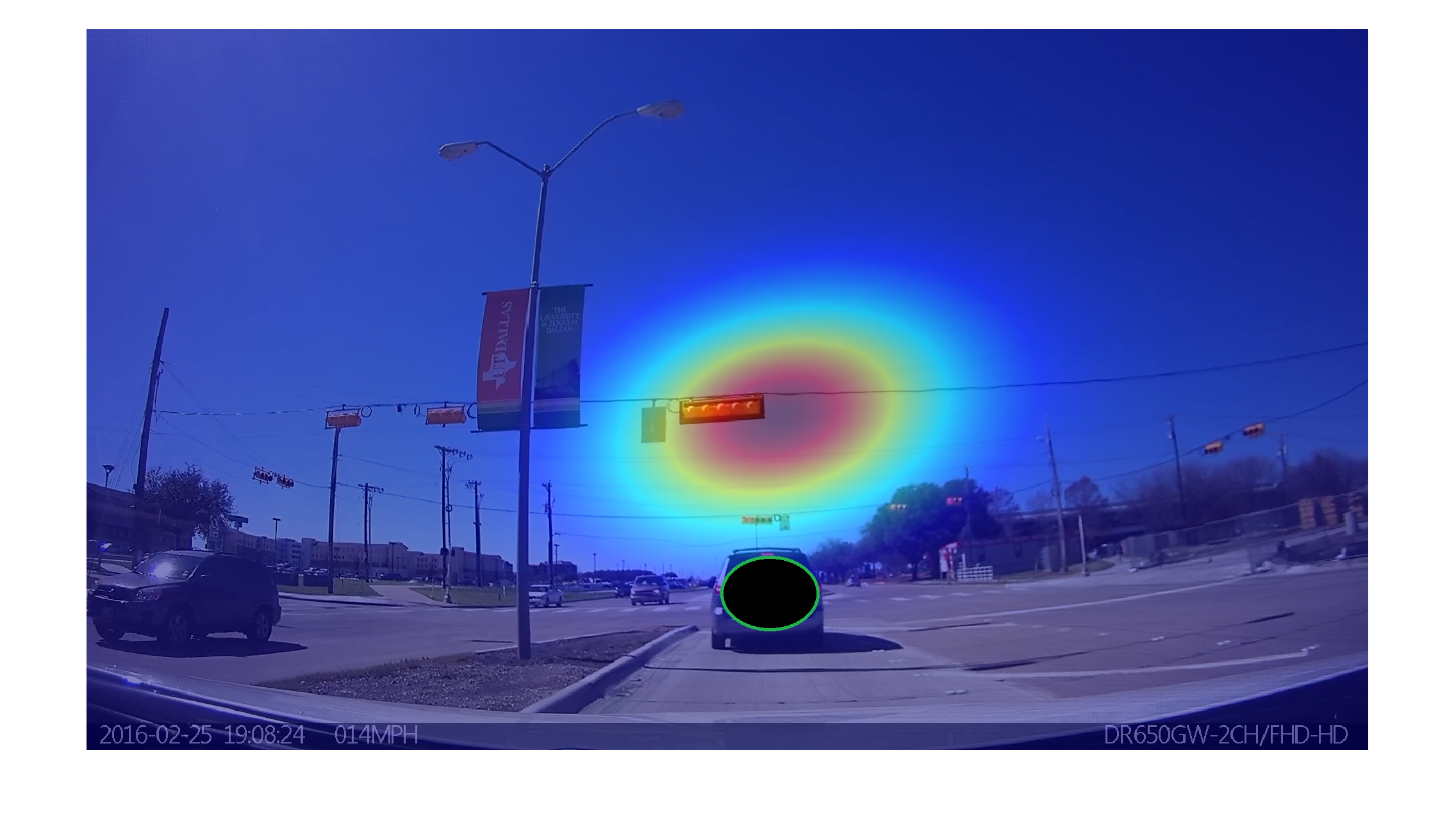}
\label{fig:mappingroad_g2}
}\\
\subfigure[Face camera]
{
\includegraphics[height=2.4cm,trim = 15cm 7cm 11cm 0cm, clip,trim = 15cm 5cm 11cm 2cm, clip]{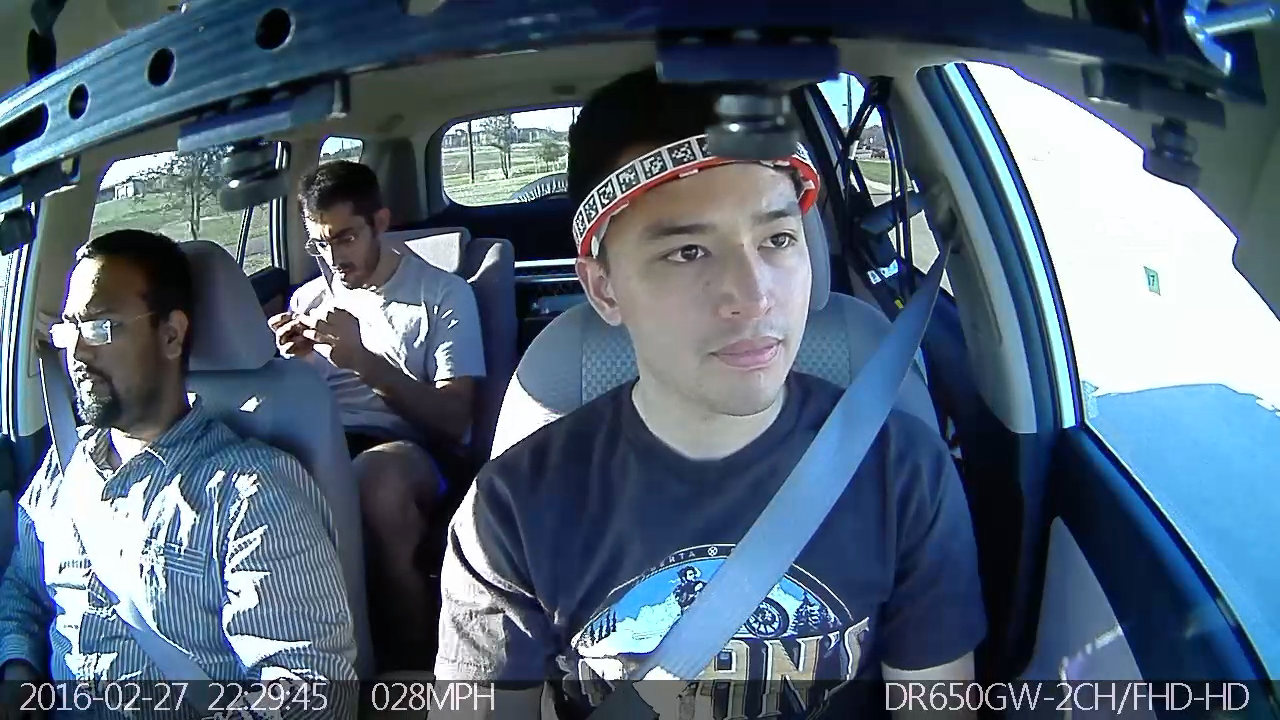}
\label{fig:mappingroad_f3}
}\hspace{-0.3cm}
\subfigure[Mapping on road, GPR]
{
\includegraphics[height=2.4cm,trim = 2cm 2cm 2cm 2cm, clip]{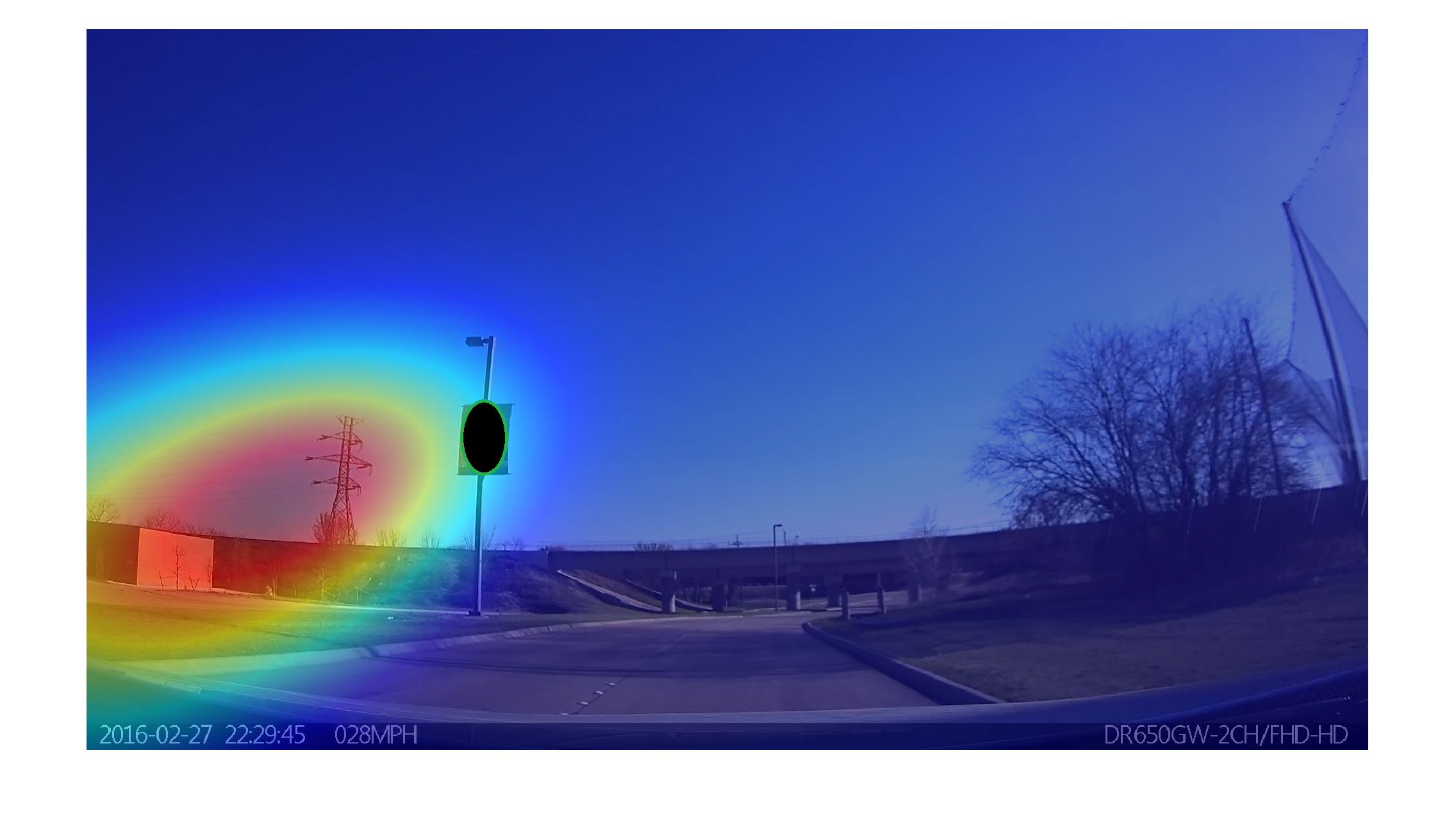}
\label{fig:mappingroad_g3}
}\\
\caption{Three examples of projections of the probabilistic salient visual maps onto the road. These regions are estimated at different distance, combining the results into a single probabilistic map. The target marker is highlighted with a black ellipse.}
\label{fig:mappingroad}
\end{figure}

\subsection{Mapping Confidence Regions onto the Road}
\label{ssec:ResultsRoad}

We also projected the predicted confidence regions onto the road view. As explained in Section \ref{ssec:Protocol}, we asked three of the subjects to look at multiple targets on the road. For these cases, we approximate the gaze distribution in the road by projecting the confident regions at different distances from the car, ranging from 10 to 200 meters in increments of 10 meters (i.e., 20 different projections). Then, we calculated the unweighted average of the probabilities for each pixel creating a 2D visual map projected on the road camera. 

Figure \ref{fig:mappingroad} gives three examples, showing the driver’s face, the road view, and the estimated salient visual map created with the GPR models. The target object is highlighted with a black ellipse. We observe that the GPR models perform reasonably well providing an prediction around the target regions attracting the attention of the driver. Notice that in this study we only consider the position and orientation of the head.

 We observe that the predicted probabilistic salient visual maps do not always include the true gaze target. These cases are useful to identify some limitations of our model to project the region on the road. First, we add some distortion during the projections, as discussed before.  Second, some subjects may depend on subtle eye movements that our models do not capture. Notice that the eye information is not used by our models, which is used in most of the gaze detection system designed for HCI in controlled environment. Third, the inter-driver variability can impact the results, as differences in height and driving behaviors can affect the relationship between head movements and gaze. In spite of these limitations, the results in this paper demonstrate that our models effectively capture the visual attention of the drivers by just modeling their head pose. We include a video with the results as a supplemental document.

\subsection{Predicting Gaze Angle with Limited Head Pose Information}
\label{ssec:ResultsLimited}

RGB cameras are the most common sensors that are used to capture the driver data in the car. Since regular cameras lack depth information, it is not possible for algorithms to reliably estimate the head position in all three degrees of freedom. Our GPR models require this information to estimate the probabilistic salient visual maps. Therefore, we retrain our GPR models by using only head orientation, or by augmenting head orientation with partial head position. We consider two conditions. The first condition only considers the head orientation (i.e., 3D vector). The second conditions is head orientation plus the $x$ and $y$ position of the head, estimated with the AprilTag-based headband. These models are compared with the GPR models trained with the 6D vector, including full orientation and position of the head.

\begin{figure}[!t]
\centering
\subfigure[GPR Linear]{
    \includegraphics[width=0.98\columnwidth,trim = 5.0cm 0cm 5.5cm 0cm, clip]{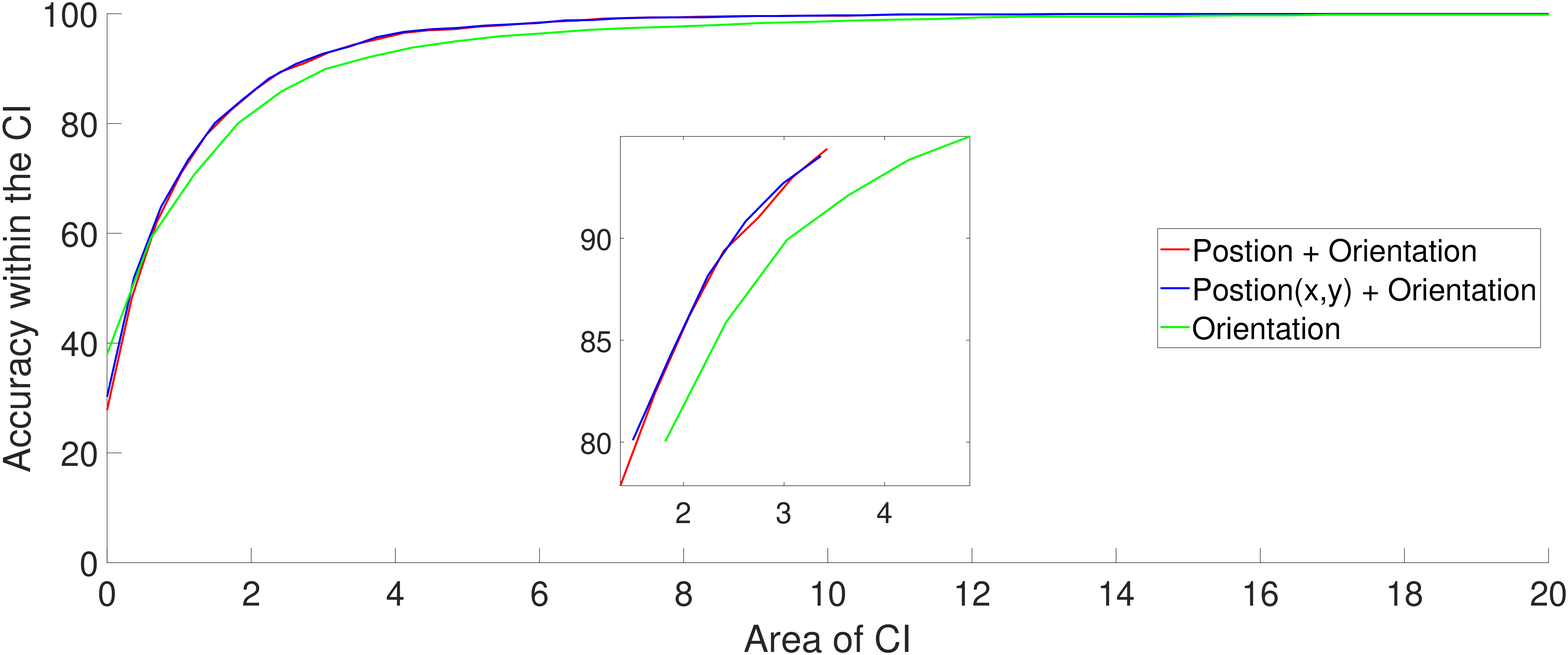}    
    \label{fig:lim_gpr_a}
}
\subfigure[GPR NN]{
    \includegraphics[width=0.98\columnwidth,trim = 5.0cm 0cm 5.5cm 0cm, clip]{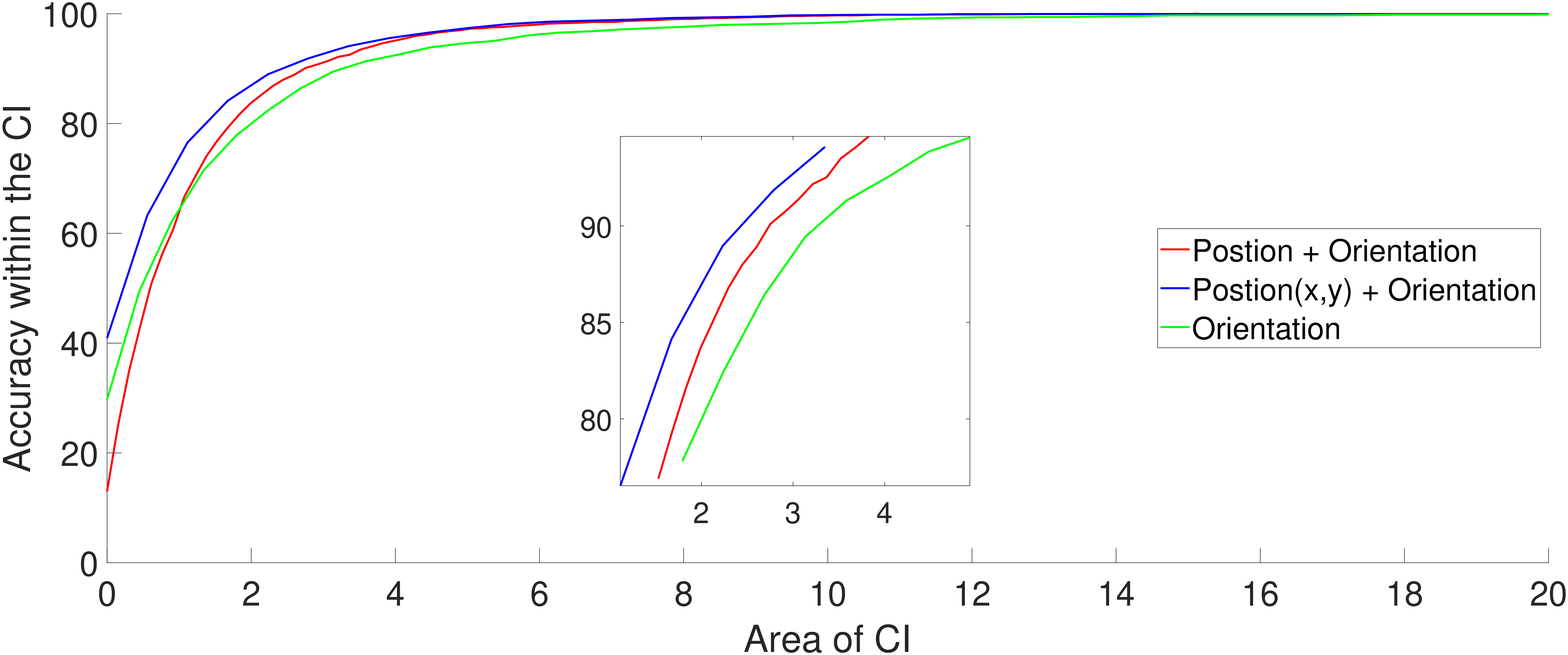}
    \label{fig:lim_gpr_b}
}
\caption{Comparison of the accuracy versus temporal resolution of GPR models implemented with limited head pose information. We zoom the plots for better visualization.  The results are separately reported for GPR linear and GPR NN. The figure is better viewed in colors.}
\label{fig:lim_gpr}
\end{figure}

Figure \ref{fig:lim_gpr} presents the results for the GPR models on the test set in phase 2 (driving condition). The GPR linear model with orientation and partial position information achieves results that are very close to the results achieved by the full model (Fig. \ref{fig:lim_gpr_a}). The GPR NN model implemented with partial head position even outperforms the results of the model with full information (Fig. \ref{fig:lim_gpr_b}). We conclude that the distance between the driver and the camera is not critical to build an effective model. We hypothesize that this results is due to the reduced head movements along the z-direction observed while a driver is operating a vehicle. The performances of both models drop when they are exclusively trained with head orientation, indicating that some information about the head position is needed.

\section{Conclusions}
\label{sec:concl}

This paper proposed a novel probabilistic model based on GPR to define a salient visual map to predict the driver's gaze location. The proposed method predicts confidence regions containing the gaze direction of the driver using only the position and orientation of his/her head. The size of the confidence region is determined by the uncertainty of the model in the predicted region (heteroscedastic model). To demonstrate the potential use of the proposed method, we projected this salient visual map onto the windshield and the road images. The results demonstrated reasonable performance suggesting that the proposed approach can play an important role in vehicle applications for security, infotainment, and navigation. The study relies on a commercial dash camera that can be easily installed on regular vehicles. The device supports WiFi connection providing a perfect platform for real-time implementation.

The paper opens new areas of research. Once the salient visual map is created and projected onto the road scene, we can leverage computer vision algorithms to detect target objects within the highlighted area. We can look more closely at the region and predict what possible objects the driver is directing his/her gaze (other vehicles, pedestrians, or billboards). We can also determine important objects that a driver fails to look at, creating an appropriate warning. For example, an ADAS can identify cases where the driver is performing a maneuver without paying attention to other vehicles. The salient visual map can also be helpful in designing smart infotainment systems. We can incorporate this information to resolve ambiguous queries, reducing the cognitive load of the driver.

There are open challenges to accurately predict the six degrees of freedom for the driver's head in real driving conditions. We use AprilTags for this purpose in our analysis. Using a single RGB camera, it can be difficult to track the head pose in a vehicle when the rotation is higher than a given threshold (e.g., when the face is not completely visible \cite{Jha_2017_2}). We also need to predict the distance of the driver's head from the camera, which will affect the gaze angle. To address this challenge, we are working on using depth sensors to reliably estimate the orientation and position of the head \cite{Hu_2020}. One of the limitations of the head band used in this study is that it covers parts of the face. We are working on an alternative design that addresses this limitation \cite{Jha_2018}. This type of data collection protocol can serve as valuable resources to train and evaluate head pose algorithms in real driving scenarios, advancing algorithm development in this area.


%
\ifCLASSOPTIONcompsoc
  \section*{Acknowledgments}
\else
  \section*{Acknowledgment}
\fi

This work was supported by Semiconductor Research Corporation (SRC) Texas Analog Center of Excellence (TxACE), under task 2810.014.


%




\ifCLASSOPTIONcaptionsoff
  \newpage
\fi



\bibliographystyle{IEEEtran}
\bibliography{reference}

\begin{thebibliography}{10}
\providecommand{\url}[1]{#1}
\csname url@samestyle\endcsname
\providecommand{\newblock}{\relax}
\providecommand{\bibinfo}[2]{#2}
\providecommand{\BIBentrySTDinterwordspacing}{\spaceskip=0pt\relax}
\providecommand{\BIBentryALTinterwordstretchfactor}{4}
\providecommand{\BIBentryALTinterwordspacing}{\spaceskip=\fontdimen2\font plus
\BIBentryALTinterwordstretchfactor\fontdimen3\font minus
  \fontdimen4\font\relax}
\providecommand{\BIBforeignlanguage}[2]{{%
\expandafter\ifx\csname l@#1\endcsname\relax
\typeout{** WARNING: IEEEtran.bst: No hyphenation pattern has been}%
\typeout{** loaded for the language `#1'. Using the pattern for}%
\typeout{** the default language instead.}%
\else
\language=\csname l@#1\endcsname
\fi
#2}}
\providecommand{\BIBdecl}{\relax}
\BIBdecl

\bibitem{Klauer_2006}
S.~Klauer, T.~Dingus, V.~Neale, J.~Sudweeks, and D.~Ramsey, ``The impact of
  driver inattention on near-crash/crash risk: An analysis using the 100-car
  naturalistic driving study data,'' National Highway Traffic Safety
  Administration, Blacksburg, VA, USA, Technical Report DOT HS 810 594, April
  2006.

\bibitem{Misu_2015}
T.~Misu, ``Visual saliency and crowdsourcing-based priors for an in-car
  situated dialog system,'' in \emph{International conference on Multimodal
  interaction (ICMI 2015)}, Seattle, WA, USA, November 2015, pp. 75--82.

\bibitem{Li_2015}
N.~Li and C.~Busso, ``Predicting perceived visual and cognitive distractions of
  drivers with multimodal features,'' \emph{IEEE Transactions on Intelligent
  Transportation Systems}, vol.~16, no.~1, pp. 51--65, February 2015.

\bibitem{Koma_2017}
H.~Koma, T.~Harada, A.~Yoshizawa, and H.~Iwasaki, ``Detecting cognitive
  distraction using random forest by considering eye movement type,''
  \emph{International Journal of Cognitive Informatics and Natural Intelligence
  (IJCINI)}, vol.~11, no.~1, pp. 16--28, January-March 2017.

\bibitem{Li_2016}
N.~Li and C.~Busso, ``Detecting drivers' mirror-checking actions and its
  application to maneuver and secondary task recognition,'' \emph{IEEE
  Transactions on Intelligent Transportation Systems}, vol.~17, no.~4, pp.
  980--992, April 2016.

\bibitem{Doshi_2009_2}
A.~Doshi and M.~Trivedi, ``Investigating the relationships between gaze
  patterns, dynamic vehicle surround analysis, and driver intentions,'' in
  \emph{IEEE Intelligent Vehicles Symposium (IV 2009)}, Xi'an, China, July
  2009, pp. 887--892.

\bibitem{Wang_2019}
Z.~Wang, R.~Zheng, T.~Kaizuka, and K.~Nakano, ``Relationship between gaze
  behavior and steering performance for driver-automation shared control: A
  driving simulator study,'' \emph{IEEE Transactions on Intelligent Vehicles},
  vol.~4, no.~1, pp. 154--166, March 2019.

\bibitem{Li_2013}
N.~Li, J.~Jain, and C.~Busso, ``Modeling of driver behavior in real world
  scenarios using multiple noninvasive sensors,'' \emph{IEEE Transactions on
  Multimedia}, vol.~15, no.~5, pp. 1213--1225, August 2013.

\bibitem{Ahlstrom_2013}
C.~Ahlstrom, K.~Kircher, and A.~Kircher, ``A gaze-based driver distraction
  warning system and its effect on visual behavior,'' \emph{IEEE Transactions
  on Intelligent Transportation Systems}, vol.~14, no.~2, pp. 965--973, June
  2013.

\bibitem{Li_2018}
N.~Li and C.~Busso, ``Calibration free, user independent gaze estimation with
  tensor analysis,'' \emph{Image and Vision Computing}, vol.~74, pp. 10--20,
  June 2018.

\bibitem{Baluja_1994}
S.~Baluja and D.~Pomerleau, ``Non-intrusive gaze tracking using artificial
  neural networks,'' Carnegie Mellon University, Pittsburgh, PA, USA, Tech.
  Rep. CMU-CS-94-102, January 1994.

\bibitem{Jha_2019}
S.~Jha and C.~Busso, ``Estimation of gaze region using two dimensional
  probabilistic maps constructed using convolutional neural networks,'' in
  \emph{IEEE International Conference on Acoustics, Speech and Signal
  Processing (ICASSP 2019)}, Brighton, UK, May 2019, pp. 3792--3796.

\bibitem{Jha_2017_2}
------, ``Challenges in head pose estimation of drivers in naturalistic
  recordings using existing tools,'' in \emph{IEEE International Conference on
  Intelligent Transportation (ITSC)}, Yokohama, Japan, October 2017, pp.
  1624--1629.

\bibitem{Tawari_2014_2}
A.~Tawari and M.~Trivedi, ``Robust and continuous estimation of driver gaze
  zone by dynamic analysis of multiple face videos,'' in \emph{IEEE Intelligent
  Vehicles Symposium (IV 2014)}, Dearborn, MI, June 2014, pp. 344--349.

\bibitem{Lee_2011_2}
S.~J. Lee, J.~Jo, H.~G. Jung, K.~R. Park, and J.~Kim, ``Real-time gaze
  estimator based on driver's head orientation for forward collision warning
  system,'' \emph{IEEE Transactions on Intelligent Transportation Systems},
  vol.~12, no.~1, pp. 254--267, March 2011.

\bibitem{Chuang_2014}
M.~C. Chuang, R.~Bala, E.~A. Bernal, P.~Paul, and A.~Burry, ``Estimating gaze
  direction of vehicle drivers using a smartphone camera,'' in \emph{EEE
  Conference on Computer Vision and Pattern Recognition Workshops (CVPRW
  2014)}, Columbus, OH, USA, June 2014, pp. 165--170.

\bibitem{Rezaei_2014}
M.~Rezaei and R.~Klette, ``Look at the driver, look at the road: No
  distraction! no accident!'' in \emph{IEEE Conference on Computer Vision and
  Pattern Recognition (CVPR 2014)}, Columbus, OH, June 2014, pp. 129--136.

\bibitem{Jha_2016}
S.~Jha and C.~Busso, ``Analyzing the relationship between head pose and gaze to
  model driver visual attention,'' in \emph{IEEE International Conference on
  Intelligent Transportation Systems (ITSC 2016)}, Rio de Janeiro, Brazil,
  November 2016, pp. 2157--2162.

\bibitem{Jha_2017}
------, ``Probabilistic estimation of the driver's gaze from head orientation
  and position,'' in \emph{IEEE International Conference on Intelligent
  Transportation (ITSC)}, Yokohama, Japan, October 2017, pp. 1630--1635.

\bibitem{Jha_2018_2}
------, ``Probabilistic estimation of the gaze region of the driver using dense
  classification,'' in \emph{IEEE International Conference on Intelligent
  Transportation (ITSC 2018)}, Maui, HI, USA, November 2018, pp. 697--702.

\bibitem{Liang_2010}
Y.~Liang and J.~Lee, ``Combining cognitive and visual distraction: Less than
  the sum of its parts,'' \emph{Accident Analysis \& Prevention}, vol.~42,
  no.~3, pp. 881--890, May 2010.

\bibitem{Robinson_1972}
G.~H. Robinson, D.~J. Erickson, G.~L. Thurston, and R.~L. Clark, ``Visual
  search by automobile drivers,'' \emph{Human Factors: The Journal of the Human
  Factors and Ergonomics Society}, vol.~14, no.~4, pp. 315--323, August 1972.

\bibitem{Underwood_2003}
G.~Underwood, P.~Chapman, N.~Brocklehurst, J.~Underwood, and D.~Crundall,
  ``Visual attention while driving: sequences of eye fixations made by
  experienced and novice drivers,'' \emph{Ergonomics}, vol.~46, no.~6, pp.
  629--646, May 2003.

\bibitem{Sodhi_2002}
M.~Sodhi, B.~Reimer, and I.~Llamazares, ``Glance analysis of driver eye
  movements to evaluate distraction,'' \emph{Behavior Research Methods,
  Instruments, \& Computers}, vol.~34, no.~4, pp. 529--538, November 2002.

\bibitem{Kutila_2007}
M.~Kutila, M.~Jokela, G.~Markkula, and M.~Rue, ``Driver distraction detection
  with a camera vision system,'' in \emph{IEEE International Conference on
  Image Processing (ICIP 2007)}, vol.~6, San Antonio, Texas, USA, September
  2007, pp. 201--204.

\bibitem{Liang_2007}
Y.~Liang, M.~Reyes, and J.~Lee, ``Real-time detection of driver cognitive
  distraction using support vector machines,'' \emph{IEEE Transactions on
  Intelligent Transportation Systems}, vol.~8, no.~2, pp. 340--350, June 2007.

\bibitem{Murphy_2008}
E.~Murphy-Chutorian and M.~Trivedi, ``{HyHOPE}: Hybrid head orientation and
  position estimation for vision-based driver head tracking,'' in \emph{IEEE
  Intelligent Vehicles Symposium (IV 2008)}, Eindhoven, The Netherlands, June
  2008, pp. 512--517.

\bibitem{Alletto_2016}
S.~Alletto, A.~Palazzi, F.~Solera, S.~Calderara, and R.~Cucchiara,
  ``{DR(eye)VE}: A dataset for attention-based tasks with applications to
  autonomous and assisted driving,'' in \emph{IEEE Conference on Computer
  Vision and Pattern Recognition Workshops (CVPRW 2016)}, Las Vegas, NV, USA,
  June-July 2016, pp. 54--60.

\bibitem{Bojarski_2016}
M.~Bojarski, D.~{Del Testa}, D.~Dworakowski, B.~Firner, B.~Flepp, P.~Goyal,
  L.~Jackel, M.~Monfort, U.~Muller, J.~Zhang, X.~Zhang, J.~Zhao, and K.~Zieba,
  ``End to end learning for self-driving cars,'' \emph{ArXiv e-prints
  (arXiv:1604.07316)}, pp. 1--9, April 2016.

\bibitem{Zeeb_2016}
K.~Zeeb, A.~Buchner, and M.~Schrauf, ``Is take-over time all that matters? the
  impact of visual-cognitive load on driver take-over quality after
  conditionally automated driving,'' \emph{Accident Analysis \& Prevention},
  vol.~92, pp. 230--239, July 2016.

\bibitem{Vora_2017}
S.~Vora, A.~Rangesh, and M.~M. Trivedi, ``On generalizing driver gaze zone
  estimation using convolutional neural networks,'' in \emph{IEEE Intelligent
  Vehicles Symposium (IV 2017)}, Los Angeles, CA, USA, June 2017, pp. 849--854.

\bibitem{Angkititrakul_2007_2}
P.~Angkititrakul, M.~Petracca, A.~Sathyanarayana, and J.~Hansen, ``{UTD}rive:
  Driver behavior and speech interactive systems for in-vehicle environments,''
  in \emph{IEEE Intelligent Vehicles Symposium}, Istanbul, Turkey, June 2007,
  pp. 566--569.

\bibitem{Angkititrakul_2007}
P.~Angkititrakul, D.~Kwak, S.~Choi, J.~Kim, A.~Phucphan, A.~Sathyanarayana, and
  J.~Hansen, ``Getting start with {UTD}rive: Driver-behavior modeling and
  assessment of distraction for in-vehicle speech systems,'' in
  \emph{Interspeech 2007}, Antwerp, Belgium, August 2007, pp. 1334--1337.

\bibitem{Hansen_2017}
J.~Hansen, C.~Busso, Y.~Zheng, and A.~Sathyanarayana, ``Driver modeling for
  detection and assessment of driver distraction: Examples from the {UTDrive}
  test bed,'' \emph{IEEE Signal Processing Magazine}, vol.~34, no.~4, pp.
  130--142, July 2017.

\bibitem{Jain_2011}
J.~Jain and C.~Busso, ``Analysis of driver behaviors during common tasks using
  frontal video camera and {CAN-Bus} information,'' in \emph{IEEE International
  Conference on Multimedia and Expo (ICME 2011)}, Barcelona, Spain, July 2011.

\bibitem{Li_2013_2}
N.~Li and C.~Busso, ``Evaluating the robustness of an appearance-based gaze
  estimation method for multimodal interfaces,'' in \emph{International
  conference on multimodal interaction (ICMI 2013)}, Sydney, Australia,
  December 2013, pp. 91--98.

\bibitem{Hu_2020}
T.~Hu, S.~Jha, and C.~Busso, ``Robust driver head pose estimation in
  naturalistic conditions from point-cloud data,'' in \emph{IEEE Intelligent
  Vehicles Symposium (IV2020)}, Las Vegas, NV USA, June 2020.

\bibitem{olson_2011}
E.~Olson, ``{AprilTag}: A robust and flexible visual fiducial system,'' in
  \emph{IEEE International Conference on Robotics and Automation (ICRA 2011)},
  Shanghai, China, May 2011, pp. 3400--3407.

\bibitem{Baltrusaitis_2018}
T.~Baltru{\v{s}}aitis, A.~Zadeh, Y.~C. Lim, and L.~Morency, ``{OpenFace 2.0}:
  Facial behavior analysis toolkit,'' in \emph{IEEE Conference on Automatic
  Face and Gesture Recognition (FG 2018)}, Xi'an, China, May 2018, pp. 59--66.

\bibitem{kabsch_1978}
W.~Kabsch, ``A discussion of the solution for the best rotation to relate two
  sets of vectors,'' \emph{Acta Crystallographica Section A.}, vol. A34, no.
  Part 5, pp. 827--828, September 1978.

\bibitem{Rasmussen_2004}
C.~E. Rasmussen, ``Gaussian processes in machine learning,'' in \emph{Advanced
  Lectures on Machine Learning}, ser. Lecture Notes in Computer Science,
  O.~Bousquet, U.~{von Luxburg}, and G.~R\"{a}tsch, Eds.\hskip 1em plus 0.5em
  minus 0.4em\relax Berlin, German: Springer Berlin Heidelberg, October 2004,
  pp. 63--71.

\bibitem{Bishop_1994}
\BIBentryALTinterwordspacing
C.~Bishop, ``Mixture density networks,'' Aston University, Birmingham, UK,
  Technical Report NCRG/94/004, February 1994. [Online]. Available:
  \url{http://www.ncrg.aston.ac.uk/}
\BIBentrySTDinterwordspacing

\bibitem{Kingma_2014_2}
D.~Kingma and J.~Ba, ``Adam: A method for stochastic optimization,'' in
  \emph{International Conference on Learning Representations}, San Diego, CA,
  USA, May 2015, pp. 1--13.

\bibitem{Chollet_2015}
\BIBentryALTinterwordspacing
F.~Chollet, ``{Keras}: Deep learning library for {Theano} and {TensorFlow},''
  https://keras.io/, April 2017. [Online]. Available:
  \url{https://github.com/fchollet/keras}
\BIBentrySTDinterwordspacing

\bibitem{Abadi_2016}
M.~Abadi, P.~Barham, J.~Chen, Z.~Chen, A.~Davis, J.~Dean, M.~Devin,
  S.~Ghemawat, G.~Irving, M.~Isard, M.~Kudlur, J.~Levenberg, R.~Monga,
  S.~Moore, D.~Murray, B.~Steiner, P.~Tucker, V.~Vasudevan, P.~Warden,
  M.~Wicke, Y.~Yu, and X.~Zheng, ``{TensorFlow}: A system for large-scale
  machine learning,'' in \emph{Symposium on Operating Systems Design and
  Implementation (OSDI 2016)}, Savannah, GA, USA, November 2016, pp. 265--283.

\bibitem{Jha_2018}
S.~Jha and C.~Busso, ``{Fi-Cap}: Robust framework to benchmark head pose
  estimation in challenging environments,'' in \emph{IEEE International
  Conference on Multimedia and Expo (ICME 2018)}, San Diego, CA, USA, July
  2018, pp. 1--6.

\end{thebibliography}

%

\begin{IEEEbiography}[{\includegraphics[width=1in,height=1.25in,clip,keepaspectratio]{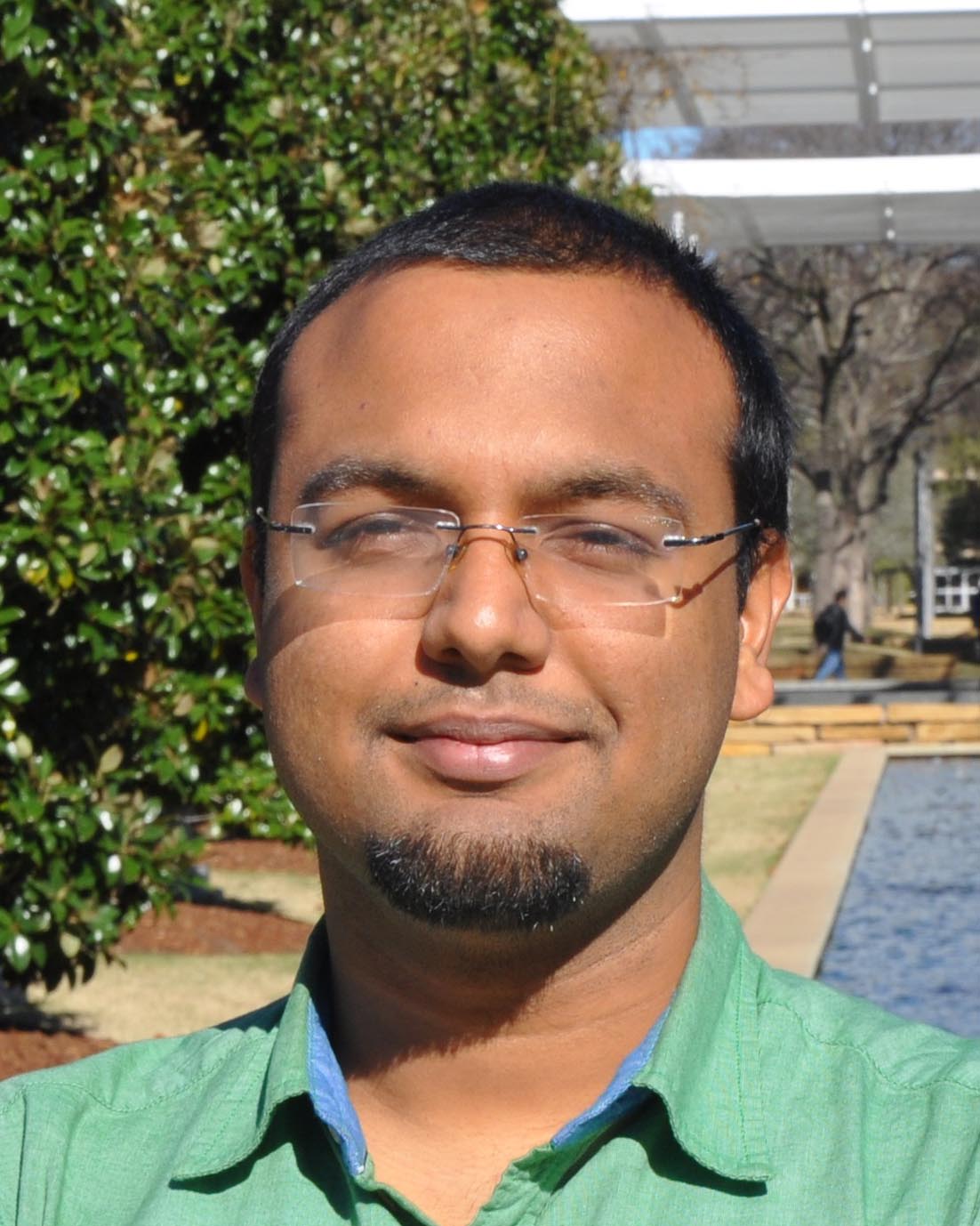}}]{Sumit Jha}
(S'16) received the B.Tech degree in Electronics and Communication Engineering form the National Institute of Technology(NIT), Trichy, India, in 2012 and the MS degree in Electrical Engineering from the University of Texas at Dallas (UTD), Texas in 2016. He is currently a PhD candidate at the University of Texas at Dallas. At UTD, he has been a part of the Multimodal Signal Processing(MSP) laboratory since 2015. His research interest lies in machine learning computer vision solutions for driver monitoring and in-vehicle safety systems. He has been a student member of IEEE since 2016.
\end{IEEEbiography}


\begin{IEEEbiography}[{\includegraphics[width=1in,height=1.25in,clip,keepaspectratio]{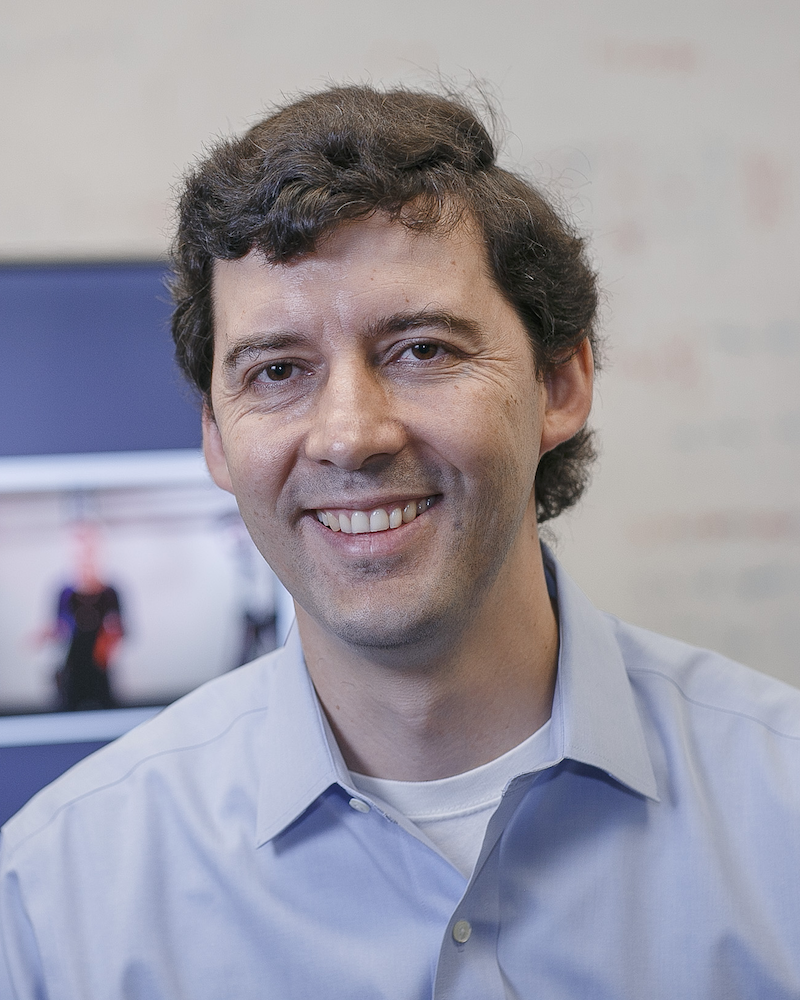}}]{Carlos Busso} 
(S'02-M'09-SM'13) received the BS and MS degrees with high honors in electrical engineering from the University of Chile, Santiago, Chile, in 2000 and 2003, respectively, and the PhD degree (2008) in electrical engineering from the University of Southern California (USC), Los Angeles, in 2008. He is an associate professor at the Electrical Engineering Department of The University of Texas at Dallas (UTD). He was selected by the School of Engineering of Chile as the best electrical engineer graduated in 2003 across Chilean universities. At USC, he received a provost doctoral fellowship from 2003 to 2005 and a fellowship in Digital Scholarship from 2007 to 2008. At UTD, he leads the Multimodal Signal Processing (MSP) laboratory [http://msp.utdallas.edu]. He is a recipient of an NSF CAREER Award. In 2014, he received the ICMI Ten-Year Technical Impact Award. In 2015, his student received the third prize IEEE ITSS Best Dissertation Award (N. Li). He also received the Hewlett Packard Best Paper Award at the IEEE ICME 2011 (with J. Jain), and the Best Paper Award at the AAAC ACII 2017 (with Yannakakis and Cowie). He is the co-author of the winner paper of the Classifier Sub-Challenge event at the Interspeech 2009 emotion challenge. His research interest is in human-centered multimodal machine intelligence and applications. His current research includes the broad areas of affective computing, multimodal human-machine interfaces, nonverbal behaviors for conversational agents, in-vehicle active safety system, and machine learning methods for multimodal processing. His work has direct implication in many practical domains, including national security, health care, entertainment, transportation systems, and education. He was the general chair of ACII 2017 and ICMI 2021. He is a member of ISCA, and AAAC, and a senior member of the IEEE and ACM. \end{IEEEbiography}




\end{document}